%% file: paper.tex
\documentclass[sigplan,10pt]{acmart}

\renewcommand\footnotetextcopyrightpermission[1]{}

\AtBeginDocument{%
  }

\acmYear{2025}\copyrightyear{2025}
\setcopyright{acmlicensed}
\acmConference[EuroSys '25]{Twentieth European Conference on Computer Systems}{March 30--April 3, 2025}{Rotterdam, Netherlands}
\acmBooktitle{Twentieth European Conference on Computer Systems (EuroSys '25), March 30--April 3, 2025, Rotterdam, Netherlands}
\acmDOI{10.1145/3689031.3696086}
\acmISBN{979-8-4007-1196-1/25/03}

\acmSubmissionID{627}

\input{defs}

\usepackage{etoolbox}
\makeatletter
\patchcmd{\maketitle}
 {\def\@makefnmark}
 {\def\@makefnmark{}\def\useless@macro}
 {}{}
\makeatother

\begin{document}

\title{Stateful Large Language Model Serving with \system}
\titlenote{An earlier draft of this work was published on arXiv~\cite{pensieve:arxiv} on Dec 9, 2023.}


\author{Lingfan Yu}
\orcid{0009-0008-7711-263X}
\affiliation{%
   \institution{New York University}
   \city{}
   \state{}
  \country{}
  }
\email{lingfan.yu@nyu.edu}
\author{Jinkun Lin}
\orcid{0000-0003-0035-4266}
\affiliation{%
   \institution{New York University}
   \city{}
   \state{}
   \country{}
   }
\email{jinkun.lin@nyu.edu}
\author{Jinyang Li}
\orcid{0000-0002-9574-1746}
\affiliation{%
   \institution{New York University}
   \city{}
   \state{}
   \country{}
   }
\email{jinyang@cs.nyu.edu}


\begin{abstract}
Large Language Models (LLMs) are wildly popular today and it is important to
serve them efficiently. Existing LLM serving systems are stateless across
requests. Consequently, when LLMs are used in the common setting of multi-turn
conversations, a growing log of the conversation history must be processed
alongside any request by the serving system at each turn, resulting in repeated
processing.

In this paper, we design $\system$, a system optimized for multi-turn
conversation LLM serving. $\system$ maintains the conversation state across
requests by caching previously processed history to avoid duplicate processing.
$\system$'s multi-tier caching strategy can utilize both GPU and CPU memory to
efficiently store and retrieve cached data. $\system$ also generalizes the
recent PagedAttention kernel to support attention between multiple input tokens
with a GPU cache spread over non-contiguous memory. Our evaluation shows that
$\system$ can achieve $1.14$-$3.0\times$ the throughput of \vllm and \trtllm and
significantly reduce latency.

\end{abstract}





\maketitle

\input{1_intro}

\input{2_bg}

\input{3_challenge}

\input{4_sol}

\input{5_impl}

\input{6_eval}

\input{7_rel}

\input{8_con}

\begin{acks}
We would like to thank our shepherd Yizhou Shan and the anonymous EuroSys
reviewers for their constructive feedback.
This project is partially supported by the National Science Foundation under
Grant No. 2220407 and through the NYU IT High Performance Computing
resources, services, and staff expertise.  Lingfan Yu is supported by an AMD
research award.
\end{acks}

\bibliographystyle{ACM-Reference-Format}
\bibliography{ref}


\end{document}

%% file: defs.tex
\usepackage{xspace}
\usepackage{caption}
\usepackage{subcaption}
\usepackage{multirow}
\usepackage{xcolor}
\usepackage{soul}

\ifx\nocoloredtext\undefined
    \newcommand{\textred}[1]{\textcolor{red}{#1}\xspace}
    \newcommand{\changebars}[2]{\textcolor{red}{#1}(\textcolor{blue}{\st{#2}})}
    \newcommand{\newtext}[1]{{\color{red}#1}}
\else
    \newcommand{\textred}[1]{\GenericError{}{textred is used}{}{}}
    \newcommand{\changebars}[2]{\GenericError{}{changebars is used}{}{}}
    \newcommand{\newtext}[1]{\GenericError{}{}newtext is used{}{}}
\fi

\newcommand{\system}{Pensieve\xspace}

\newcommand{\init}{prefill\xspace}
\newcommand{\gen}{generation\xspace}
\newcommand{\agen}{autoregressive generation\xspace}

\newcommand{\kvtokens}{KV-tokens\xspace}
\newcommand{\kvtoken}{KV-token\xspace}
\newcommand{\kvcache}{KV cache\xspace}
\newcommand{\pastkvtokens}{past \kvtokens}
\newcommand{\pastkvtoken}{past \kvtoken}

\newcommand{\sharegpt}{ShareGPT\xspace}
\newcommand{\ultrachat}{UltraChat\xspace}
\newcommand{\vllm}{vLLM\xspace}
\newcommand{\trtllm}{TensorRT-LLM\xspace}
\newcommand{\ORCA}{ORCA\xspace}

\newcommand{\multiattn}{multi-token attention\xspace}
\newcommand{\Multiattn}{Multi-token attention\xspace}
\newcommand{\singleattn}{single-token attention\xspace}
\newcommand{\Singleattn}{Single-token attention\xspace}
\newcommand{\gqa}{Grouped-Query Attention\xspace}

\newcommand{\opt}{OPT\xspace}
\newcommand{\llama}{Llama 2\xspace}

\newcommand{\PensievePath}{.}

\newcommand{\capsize}[1]{\small{#1}}

%% file: 1_intro.tex
\section{Introduction}
\label{pensieve:sec:intro}

The world has recently witnessed the fast expansion of Large Language Models (LLMs). The most popular use of LLM is for chatbots, with applications like ChatGPT demonstrating an astounding capability of following instructions and interacting with humans as a virtual assistant. Other LLM-backed applications include writing code, responding to emails, doing literature reviews, etc.  As LLM continues its explosive growth, it is imperative to develop fast and efficient LLM serving systems.

An LLM is an autoregressive DNN model based on the Transformer~\cite{transformer} architecture. The model iteratively predicts the next output token based on the current context which includes the sequence of input prompt tokens followed by output tokens generated in the previous iterations. LLMs require very expensive computation due to two factors. One, LLMs have huge parameter sizes (10s or 100s of billions) and have a trend of growing even larger.  Two, LLMs need to support a large context size (2K to 32K tokens) to be useful.  There has been much related work to improve the performance of LLM inference/serving from various angles, including better batching~\cite{orca}, operation fusion~\cite{flash}, better GPU memory utilization~\cite{vllm}, faster output generation~\cite{speculative:leviathan,speculative:chen}, low rank adaptation~\cite{lora} and quantization~\cite{qlora} (see \S\ref{pensieve:sec:related}).  In contrast to these works, in this paper, we take a step back and examine inefficiencies that arise in the setting of a specific but very popular LLM use case today, aka as a multi-turn conversational chatbot.

In the conversational setup, the user and the chatbot are engaged in a dialogue that may last many rounds.  In order for the chatbot not to ``lose memory'' of what has been said so far when responding, the cumulative history of the dialogue must be part of the context for LLM's autoregressive generation.  As existing LLM serving systems are stateless across requests, one must prepend a growing log of conversation history alongside each new request as the input prompt to be processed from scratch. This causes much duplicate processing for multi-turn conversations.

How to avoid duplicate processing of the chat history? To do so, the serving system can save any previously processed context data in the form of token embeddings.  When new requests from the same conversation arrive, the saved context data can be re-used and subsequently augmented.  This can be done with best effort. Essentially, the serving system is allowed to keep some cached state containing previously processed context across requests. Doing so enables the serving system to exploit the opportunity that when users are actively chatting with an AI chatbot, follow-up requests usually arrive within a reasonably short time period to leverage the cached state.

Caching state across requests is straightforward at the high level, but several challenges remain to make it really work. First, where to save the data? Keeping it in the GPU is the fastest, but is very constrained by the relatively small GPU memory size.  Putting it on disk would incur much longer load latency, hurting the user experience.  A two-tier caching solution spanning both GPU and CPU memory is promising, but care must be taken to cope with each tier's capacity limit and to swap in/out saved context data from/to the GPU without damaging performance. Second, how to reuse saved context data efficiently when processing a new request? Furthermore, some parts of the saved context can be dropped due to the cache limit. How to handle partially saved context by recomputing what has been dropped?

In this paper, we design a stateful LLM serving system, called \system, to address the aforementioned challenges.
\system saves a conversation's processed context in a two-tier GPU-CPU cache.  It evicts cached data to the next tier (or discards it), preferring conversations that have been inactive for longer and/or those that are cheaper to recompute.  The eviction is done at the granularity of a chunk of tokens instead of the whole conversation.  Therefore, a conversation's saved context might span both tiers of the cache and may be partially dropped. \system uses ahead-of-time swapping and pipelined transfer to overlap computation with the data movement between cache tiers. Dropped contexts are handled via recomputation. Evicting and restoring cause a conversation's cached context to occupy non-contiguous GPU memory.  We develop a new GPU kernel to compute attention~\cite{transformer} between multiple input tokens and cached context residing in non-contiguous memory, which is lacking in existing LLM serving systems. Our kernel is a generalized version of the PagedAttention kernel in vLLM~\cite{vllm}.

We have built \system and compared its performance against vLLM~\cite{vllm} and
\trtllm~\cite{trtllm}, two state-of-the-art serving systems that do not cache state
across requests. Experiments show that \system achieves serving throughput that is  
$1.14$-$1.70\times$ of vLLM and \trtllm for small single-GPU models (OPT-13B, Llama 2-13B) and $1.64$-$3.0\times$ for large four-GPU models (OPT-66B, Llama 2-70B).  At moderate serving load, \system also significantly reduces latency.

In summary, this paper makes the following contributions:
\begin{itemize}
\item We identify a major inefficiency of existing LLM serving systems when used for multi-turn conversations: a conversation's history context is recomputed with each successive new request in the same conversation.
\item We develop \system, a stateful LLM serving system that saves the conversation context in a multi-tier GPU-CPU cache and reuses it across requests to minimize redundant computation. Our system design can efficiently move data between cache tiers and handle partially dropped context via recomputation.
\item We build a new attention GPU kernel to compute attention between a new request's multiple input tokens and the saved context scattered in non-contiguous GPU memory. Existing kernels either require contiguous GPU context cache or are restricted to a single input token.
\item We evaluate \system using real-world conversation datasets to demonstrate its effectiveness compared to the state-of-the-art stateless serving system.
\end{itemize}

%% file: 2_bg.tex
\section{Background}
\label{pensieve:sec:bg}
\begin{figure*}[tbp]
\centering
\includegraphics[width=0.9\linewidth]{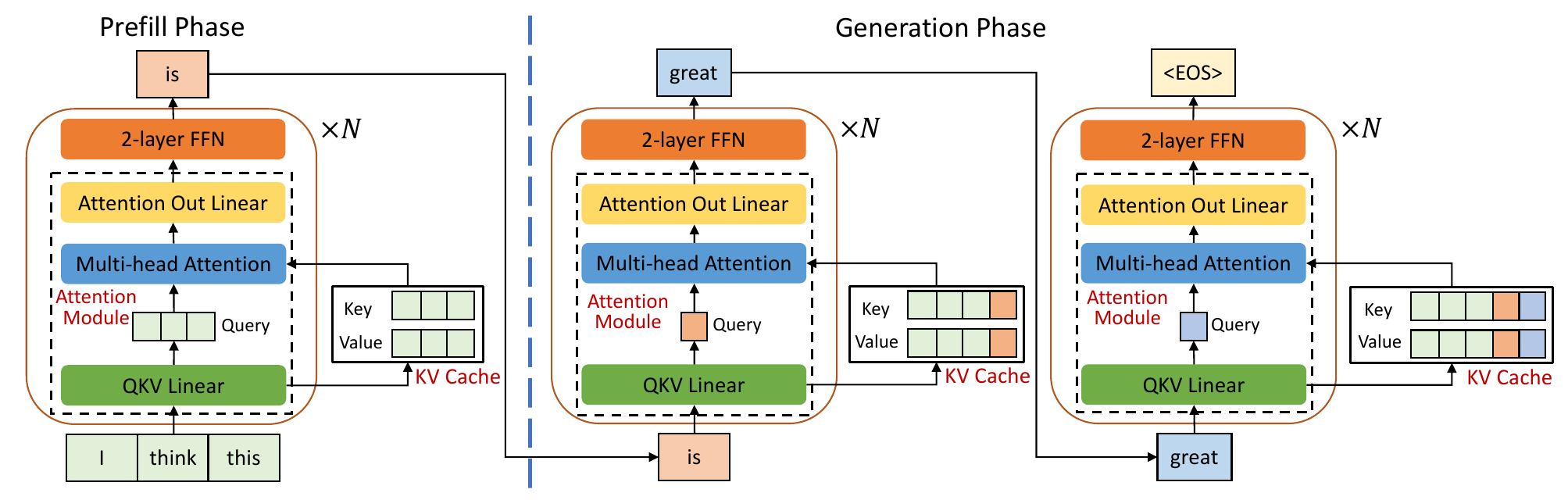}
\caption{\capsize{How inference is done for Transformer-based LLMs}}
\label{pensieve:fig:bg:llm_inference}
\end{figure*}
This section provides a brief background on LLMs and how existing systems serve them.

\subsection{LLM and the Attention Mechanism}
\label{pensieve:sec:bg:llm}
Popular large language models, e.g. GPT-3~\cite{gpt3}, OPT~\cite{opt}, Llama~\cite{llama,llama2}, are all based on the Transformer architecture~\cite{transformer}. A model consists of many transformer layers, each of which is composed of an attention module, seen as the dashed box Figure~\ref{pensieve:fig:bg:llm_inference}, and a 2-layer feed-forward network. The model takes as input a sequence of token IDs representing the natural language sentence and feeds them through an embedding layer to obtain a continuous representation (aka embedding) for each token before feeding them through transformer layers. For simplicity, we refer to token embeddings as ``tokens'', and refer to token IDs as ``raw tokens''. LLM is autoregressive in the sense that it iteratively predicts the next output token based on the current context which includes the input prompt tokens followed by output tokens generated in the previous iterations.

The success of the Transformer originates
from the capability of its attention module. For each layer, the attention module first performs QKV projection, aka linear transformations on
its input token embedding ($X$), to produce three new  embeddings Query ($Q$), Key ($K$), and Value ($V$):
\begin{align}
Q&=XW_{query}\nonumber\\
K&=XW_{key}
\label{eq:qkv_linear}\\
V&=XW_{value}\nonumber
\end{align}
where $W_{query}$, $W_{key}$ and $W_{value}$ are trainable weights.  We refer to the $K$ and $V$ embeddings as \kvtokens. The module then computes all-to-all attention scores using the dot product between each pair of tokens' query and key:
\begin{equation}
A=\frac{QK^T}{scale}
\label{eq:attn_score}
\end{equation} where scale is a normalization factor.
This attention score is then normalized using softmax and used for weighted aggregation of the token embeddings in $V$:
\begin{equation}
O=softmax(A)V
\label{eq:attn_agg}
\end{equation}
The above equations show the process of so-called single-head attention. In practice, multi-head attention is used, where $Q$, $K$, $V$
produced by Equation~\eqref{eq:qkv_linear} are divided into groups, called attention heads. Each attention head
independently performs attention as Equation~\eqref{eq:attn_score} and \eqref{eq:attn_agg}.

\subsection{How LLM is Served}
\label{pensieve:sec:bg:llm_serve}
\paragraph{The \init vs. \gen phase} To perform inference using an LLM, one needs to keep a \kvcache in GPU to avoid recomputation during the autoregressive output generation.
Figure~\ref{pensieve:fig:bg:llm_inference} shows the typical LLM inference process adopted by systems like FasterTransformer~\cite{fastertransformer}, ORCA~\cite{orca}, vLLM~\cite{vllm}. It is divided into two phases:
1) In the \init phase, all input prompt tokens are processed together to generate $K$ and $V$ (aka \kvtokens) for each layer, and the \kvcache is initialized with the resulting \kvtokens.
The embedding of the last token from the last layer is used to generate the first output token;
2) The \gen (also referred to as decoding) phase works iteratively over many steps. In each step, the token generated by the last \gen step is processed as a single new input token. Each layer computes the $Q$,$K$,$V$ embedding vector for the new token, updates the \kvcache, and performs attention using the new token's $Q$ embedding with all \kvtokens in the \kvcache.

\paragraph{Iteration-level batching} For LLMs that have variable-sized input and output, the granularity of batching has a huge impact on system throughput and serving latency. If scheduling is performed at the
request granularity, executing a batch of requests with different input prompt lengths requires padding tensors to the maximum length and waiting for the request with the longest output to finish. Iteration-level batching strategy, originally
proposed by BatchMaker~\cite{batchmaker} for non-transformer-based sequence-to-sequence models, performs batching at token granularity. ORCA~\cite{orca} extends this approach to support the LLM workload: whenever a request finishes an iterative generation step, the scheduler checks whether it has reached the end of a sequence and can leave the batch, making room for a request to start its generation phase immediately.

\paragraph{Memory management.} For each request, the model performs iterative generation until either the special end-of-sentence token (EOS) is emitted or the preconfigured maximum decoding length is reached. Systems like
FasterTransformer~\cite{fastertransformer} and ORCA~\cite{orca} reserve slots in \kvcache for each request based on the maximum decoding size. A more recent system,
vLLM~\cite{vllm}, can dynamically grow the allocated cache slots for each request and allow these slots to reside in non-contiguous GPU memory. vLLM develops PagedAttention GPU kernel to handle the \gen phase with non-contiguous \kvcache. Existing serving systems are stateless across requests. In other words, they de-allocate all the cache slots used by a request as soon as it finishes.

%% file: 3_challenge.tex
\section{Motivation and Challenges}
\label{pensieve:sec:challenge}

\subsection{Motivation}
\label{pensieve:sec:challenge:motivation}

\begin{figure}[tbp]
\centering
\includegraphics[width=\columnwidth]{\PensievePath/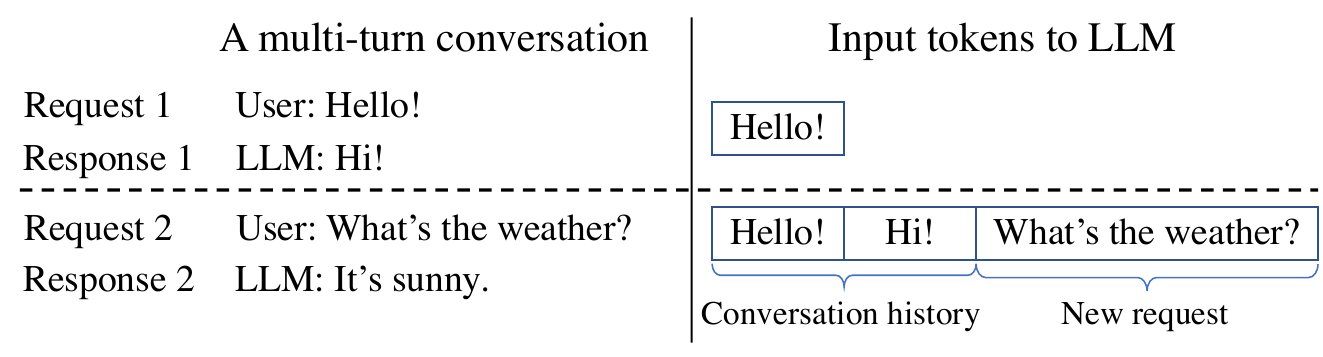}
\caption{\capsize{Existing serving systems process a cumulative history repeatedly
with each request in a multi-turn conversation.}}
\label{pensieve:fig:multiturn}
\end{figure}
Existing techniques for serving LLMs mostly focus on improving the inference time or memory efficiency of a single request. We take a step back and examine inefficiencies that arise in a very popular LLM use case today, aka as a multi-turn conversational chatbot.

In a multi-turn conversation, the user engages in multiple rounds of conversations with the chatbot so the underlying LLM needs to be aware of the conversation history to generate an appropriate response. This is done by prepending the cumulative conversation history as raw text to each new request, due to the stateless nature of existing serving systems, as shown in Figure~\ref{pensieve:fig:multiturn}. As the
interaction between the user and chatbot continues, the conversation history grows, making the cost of the \init phase overshadow that of the iterative \gen phase.  Unfortunately, much of the history processing is redundant.

\begin{figure}[t]
\centering
\includegraphics[width=0.9\columnwidth]{\PensievePath/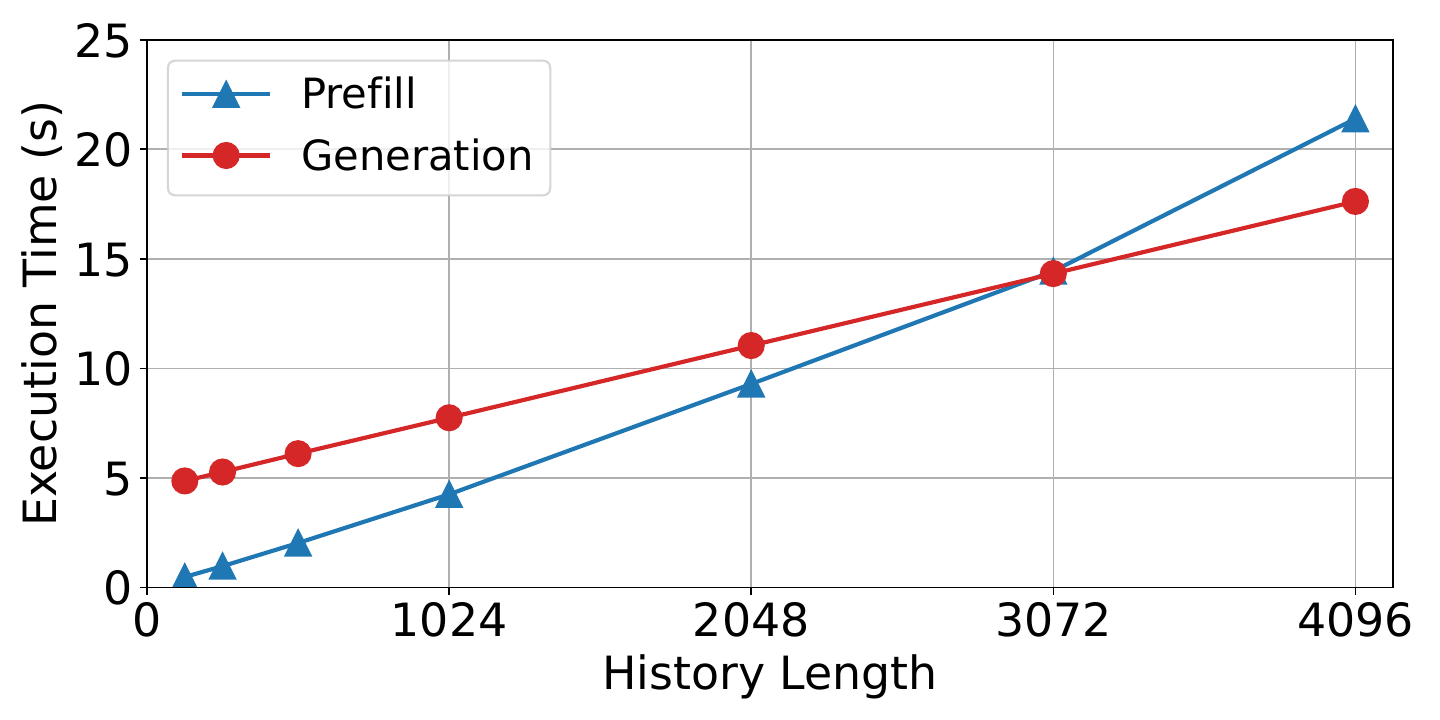}
\caption{\capsize{Execution time for a batch of 32 requests performing prompt (32
tokens) prefill and generations for 200 steps.}}
\label{pensieve:fig:challenge:init_vs_increment}
\end{figure}

Figure~\ref{pensieve:fig:challenge:init_vs_increment} demonstrates the heavy cost of the prompt initiation phase under an artificial workload where each request has 200 new prompt tokens and has varying conversation history sizes.
As shown in the figure, the cost of recomputing the conversation history (solid blue line) causes the cost of the \init phase to soon outgrow the \gen phase.

The goal of this project is to minimize redundant computation of the conversation history.  This can be done by caching any previously processed embeddings at the serving system and re-using them across requests from the same conversation.  More concretely, one can save the \kvtokens in the \kvcache belonging to a previous request and only process the new user prompts of the next follow-up request while re-using history embeddings saved in the \kvcache.

\subsection{Challenges}
\label{pensieve:sec:challenge:challenge}

\paragraph{Limited GPU memory for caching.}
LLM has very large model parameters, resulting in large \kvtokens. For
example, a 13 billion parameter GPT-3 model has 40 layers and a hidden size of 5120. Assuming the use of 16-bit half-precision numbers, storing each \kvtoken takes 2 * 40 (layer) * 5120 (units/layer) * 2 (bytes/unit) = 0.78MB memory space.
Given the limited GPU memory capacity, depending on the history lengths, only a few dozen or hundreds of conversation histories can be kept in the GPU. Therefore, we must extend our cache space to use the more abundant CPU memory.

\paragraph{Token-level cache management and recovery}
When using a multi-tier GPU-CPU cache, the serving system needs to swap cached history from GPU to CPU and vice versa.  Swapping at the coarse granularity of an entire conversation history is sub-optimal; not only does it utilize the cache space inefficiently but also it incurs large swapping latency.
Thus, we decide to swap at the granularity of individual tokens.  Specifically, in order to make room for the processing of new requests, the serving system chooses certain cached \kvtokens to swap from GPU to CPU. Later, it also needs to restore the swapped out \kvtokens from CPU to GPU.

\begin{figure}[tbp]
\centering
\includegraphics[width=0.9\columnwidth]{\PensievePath/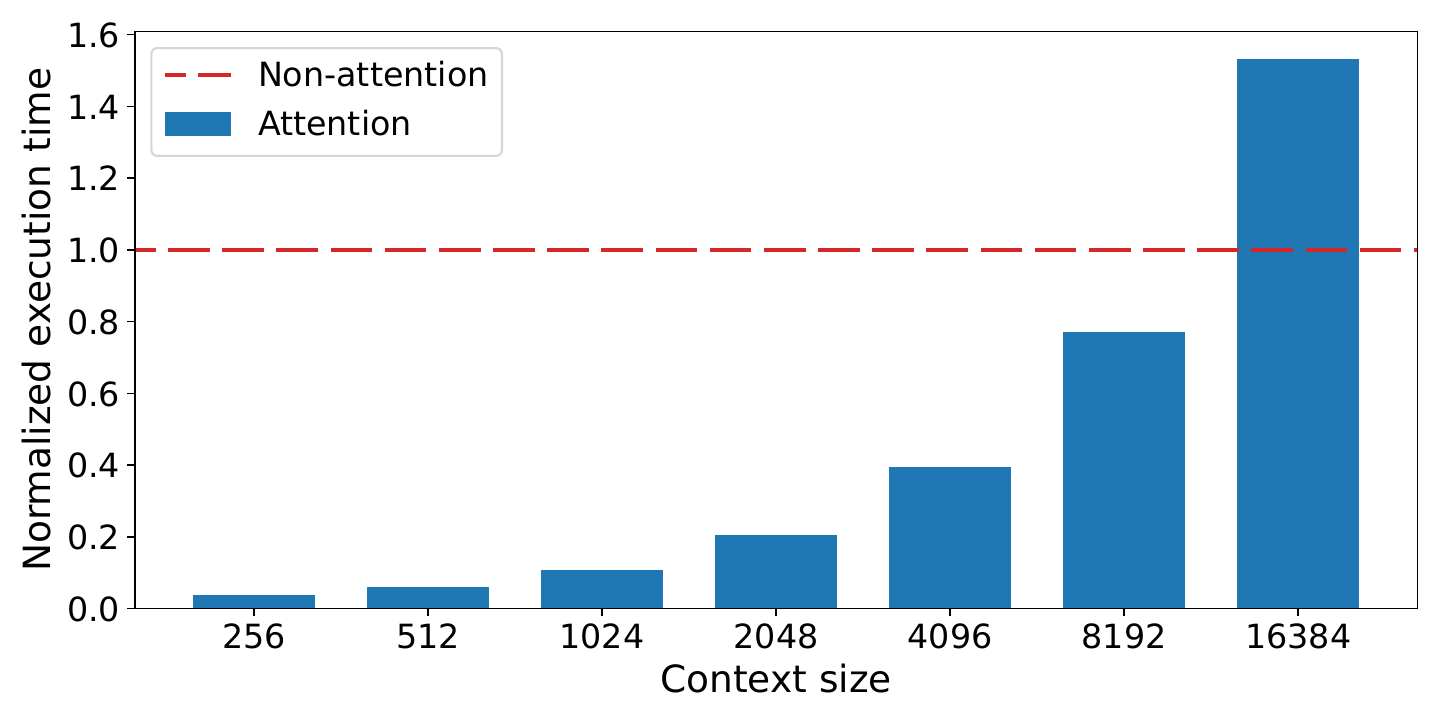}
\caption{\capsize{Execution time of attention operation for a chunk of 32 tokens
with different context sizes. Results are normalized by the execution time of
non-attention operations in a transformer layer.}}
\label{pensieve:fig:challenge:attn_cost}
\end{figure}

\begin{figure}[tbp]
\centering
\includegraphics[width=\columnwidth]{\PensievePath/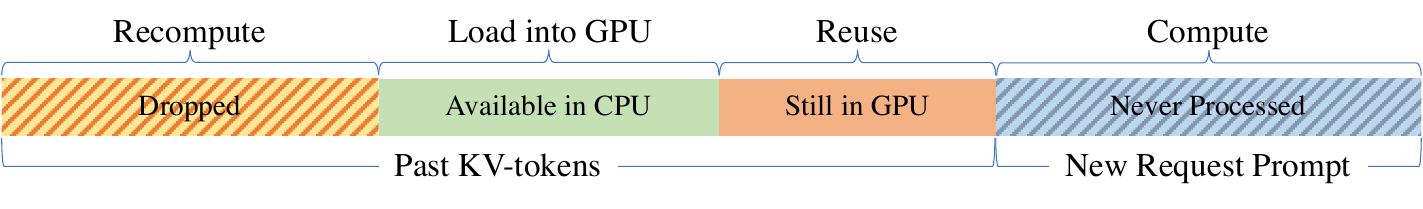}
\caption{\capsize{Layout of a typical request's \kvtoken context. The shaded
areas, which occur at both ends of the context, mark those tokens that must be
processed by the \init phase.}}
\label{pensieve:fig:challenge:recompute_drop}
\end{figure}

When the system is under CPU memory pressure, some cached \kvtokens need
to be dropped and re-computed later when needed. We note that, although each token occupies the same amount of memory space, the recomputation cost of each token is different due to the nature of causal attention computation.
Specifically, tokens appearing later in the context sequence require more computation than earlier ones because tokens ``attend to'' all preceding but not succeeding tokens.
Figure~\ref{pensieve:fig:challenge:attn_cost} shows the execution time of the
attention operator for 32 tokens with a prompt context of varying sizes; the execution time shown has been normalized against the rest of inference time (aka the sum of execution times for all other non-attention operators). As can be seen in Figure~\ref{pensieve:fig:challenge:attn_cost},
the cost of attention grows linearly with context size. Thus, when deciding which tokens to discard to reclaim CPU memory, it is more preferable to drop
the leading tokens of a conversation history.

Dropping \kvtokens from the leading end of a conversation brings additional
complexities. Figure~\ref{pensieve:fig:challenge:recompute_drop} illustrates the layout
of a typical request from a continuing conversation in its \init phase.  The request's context can be viewed as composed of four segments: 1) the first and earliest segment corresponds to tokens that have been dropped from the CPU cache and must be recomputed. 2) the second segment corresponds to tokens that reside in the CPU cache and will be fetched into the GPU. 3) the third segment contains tokens residing in the GPU cache. 4) the fourth and latest segment contains the raw tokens corresponding to this request's new prompt.
As we can see, both the first and fourth segment requires computation. However, such separation of computation at both ends of the context breaks the assumption of all existing
attention kernels that the input tokens belong in a consecutive context region in the \init phase.

\paragraph{Handling non-contiguous \kvcache.}
\begin{figure}
\centering
\includegraphics[width=\columnwidth]{\PensievePath/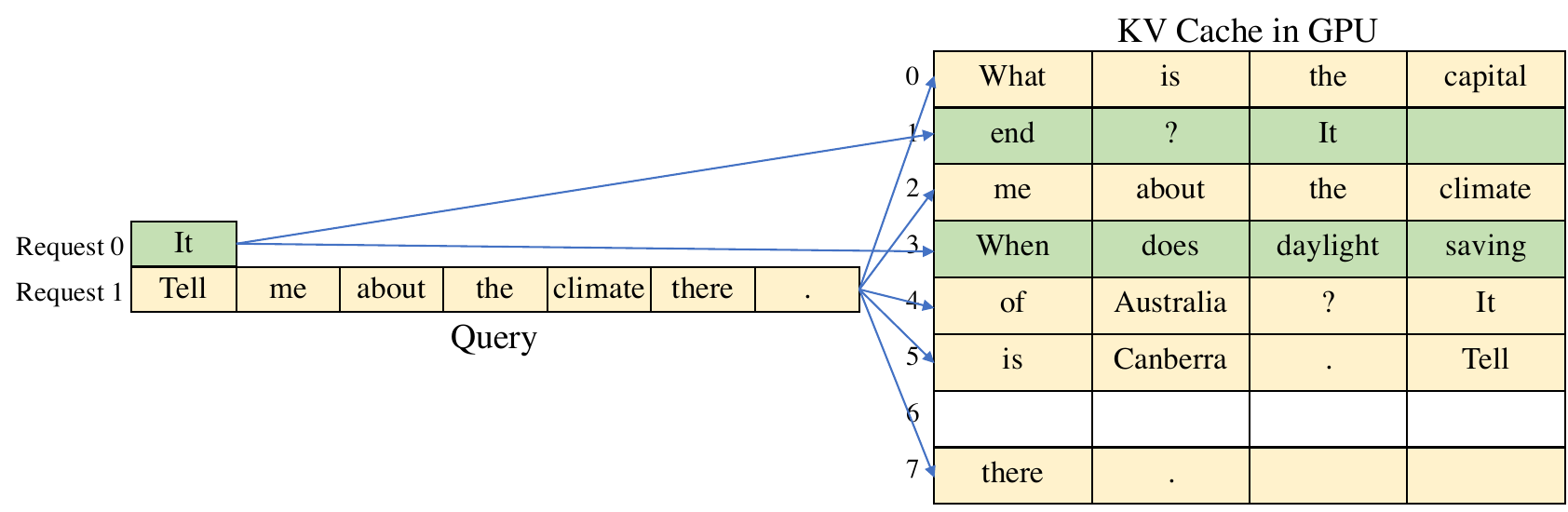}
\caption{\capsize{\Multiattn with non-contiguous \kvcache. The batch contains two
requests, whose tokens (Query representations) are shown on the left. The right
side shows the requests' context (\kvtokens) which reside in non-contiguous GPU
memory (chunk 3, 1 for request 0 and chunk 0, 4, 5, 2, 7 for request 1).}}
\label{pensieve:fig:challenge:multi_query_attn}
\end{figure}

Existing systems\xspace\cite{orca,vllm} batch requests separately for the \init and \gen phase so that they can use existing high-performance attention kernels~\cite{flash, memeff} for the \init phase.
Unfortunately, we cannot simply adopt such a design in our setting.  This is because existing attention kernels for \init assume a \kvcache with contiguous memory.  However,
in order to support the swapping of \kvtokens between GPU and CPU, it is more efficient to allow \kvtokens to reside in non-contiguous GPU memory regions.  Although vLLM~\cite{vllm} has developed the PagedAttention kernel to handle non-contiguous \kvcache, it is designed to be solely used in
the \gen phase, because it limits each request in the batch to have exactly one input token.  As each request has more than one token in the \init phase, one cannot simply use PagedAttention for \init.  A naive hack is to process the new prompt one token at a time, in the same manner as iterative \gen, so that the PagedAttention kernel can
be applied. But this method gives up the parallelization opportunity brought by the extra query token dimension in the \init phase.  Thus, to achieve efficient GPU computation, we must address the challenge of
supporting non-contiguous \kvcache during \init.  Doing so also brings extra benefit: as both the \init phase and \gen phase
can compute using the same non-contiguous \kvcache, we can handle requests in different phases together in the same batch. Figure~\ref{pensieve:fig:challenge:multi_query_attn}
illustrates the desired \Multiattn kernel for computing attention for a batch of two requests, one in its \gen phase, the other in its \init phase, over non-contiguous \kvcache.

%% file: 4_sol.tex
\section{System Design}
\label{pensieve:sec:design}

At a high level, we aim to save a conversation's \kvtokens across multiple turns in a multi-tier GPU-CPU cache.
To realize the potential performance benefits, we need to make cache swapping and dropped token recomputation efficient, by developing techniques to address
the challenges in \S\ref{pensieve:sec:challenge}.

\subsection{System Overview}
\begin{figure}[tbp]
\centering
\includegraphics[width=\columnwidth]{\PensievePath/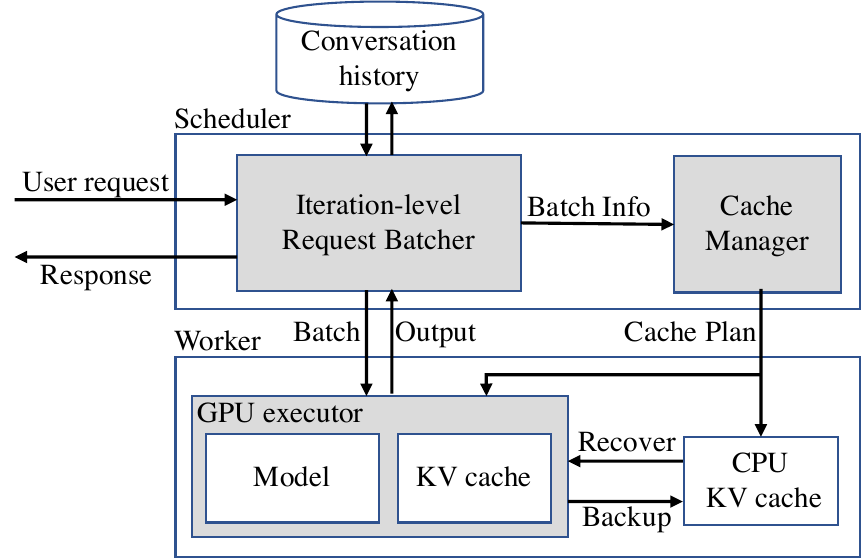}
\caption{\capsize{System architecture of \system}}
\label{pensieve:fig:system}
\end{figure}

Figure~\ref{pensieve:fig:system} shows the system architecture. \system consists of a
single scheduler and multiple workers, each of which manages one GPU.
The scheduler has two jobs: 1) it is responsible for batching requests for execution, and 2) it ensures that requests in the batch have sufficient GPU memory for execution.  For 1), the scheduler performs fine-grained iteration-level batching~\cite{batchmaker,orca,vllm} so that a new request can join the batch with existing requests while the latter is in the process of performing \agen. For 2), the scheduler manages the allocation of slots in the \kvcache and determines when to swap between the GPU and CPU \kvcache. Each worker also has two jobs: 1) it invokes GPU kernels to process a batch of requests. 2) it performs the actual data movements between the GPU and CPU \kvcache based on the batch's cache plan as determined by the scheduler.

\subsection{A unified batch scheduler}
\label{subsec:batchscheduler}

\system performs
fine-grained iteration-level batching~\cite{batchmaker, orca, vllm}. However, our batching strategy differs from existing LLM serving systems. While \ORCA only batches non-attention operations and handles attention operations individually for each request~\cite{orca}, \system and vLLM~\cite{vllm} batch requests for both non-attention and attention operations. However, unlike vLLM which only forms a batch among requests in the same phase and thus processes each of the two types of batches in separate kernel invocation, \system handles both the \init phase and \gen phase in a unified way. In other words, the \system scheduler forms a batch
of requests regardless of which phase they are in. In the same batch, some could be in their \gen phase while others are in their \init phase. Such unified batching is made possible by \system's \multiattn kernel design (\S\ref{subsubsec:unify}).

\system's iteration-level scheduler is ``clocked'' for action by the completion of a \gen step.  In particular, after each iteration of token generation, the worker returns any finished request that has either emitted the end-of-sentence token or reached maximum decoding length. The scheduler finds the next request in its wait queue to join the batch, if there is room (i.e. the total number of tokens in the batch is fewer than some pre-configured threshold).  We use the simple first-come-first-serve scheduling policy when choosing which new requests to join the batch.

For unified batch formation, the scheduler concatenates the tokens to be processed from all requests. For each new request joining the batch, its tokens include those corresponding to the request's new user prompt. For each existing request in the batch, its token includes the one generated by the last \gen step.
By combining the \init
phase together with the \gen phase, we avoid running separate small kernels and can thus improve GPU utilization.

\subsection{\kvcache management}

Traditionally, the \kvcache in GPU only serves as a computation workspace.  In
\system, the GPU \kvcache also serves as storage space to cache the \kvtokens of
recently completed requests of active conversations.  \system adopts a two-tier hierarchical caching
strategy and uses the much larger CPU memory as the second-tier cache space.
The scheduler determines the caching plan and instructs the worker to perform
the actual data movements between GPU and CPU.

The scheduler tracks the amount of free GPU \kvcache slots left.
Before handing off a batch of requests to the worker, the scheduler tries to ensure that any new request's \pastkvtokens will reside in the GPU and there is sufficient GPU memory for the execution. Specifically, if a request in the batch has any \kvtokens that have been swapped out to the CPU memory or dropped, the scheduler determines how many additional GPU \kvcache slots are needed to swap in or re-calculate those absent \kvtokens.  If there is sufficient space, the scheduler instructs the worker to perform the necessary allocation and to start swapping those \kvtokens from the CPU as part of the batch's cache plan.

\subsubsection{Eviction Policy}
\label{subsubsec:evict}
\system performs fine-grained \linebreak token-level cache eviction and dropping. We design the eviction policy to express two kinds of preferences: 1) it preferentially evicts from older conversations, aka those that have been inactive for a longer period of time.  This is based on the same LRU assumption that the least recently active conversation will not see activity for a longer time. 2) it preferentially evicts tokens from the leading end of a conversation's history context.
This is based on the observation, previously shown in Figure~\ref{pensieve:fig:challenge:attn_cost}, that leading tokens of a conversation are cheaper
to recompute than trailing ones.  Below, we describe
how our policy evicts according to these preferences.

\paragraph{Eviction granularity.}
In order to reduce the overhead caused by frequent eviction decision-making and moving small amounts of memory over the PCIe bus, we group \kvtokens into chunks and make eviction decisions at the granularity of chunks. The chunk size is configurable. In our experiments, we find that setting the chunk size to 32 tokens works well.

\paragraph{The retention value of a chunk.}
In order to combine both the LRU preference and the evicting from the front preference, we calculate a score
for each chunk to capture its retention value. Specifically,
the retention value of a chunk is $V=\frac{Cost(s, l)}{T}$, where
$Cost(s, l)$ represents the cost of recomputing a chunk of size $s$ with a
context of size $l$, and the denominator $T$ is the amount of time since the conversation was last active.  \system evicts chunks according to the ascending order of their retention values so that chunks with lower recomputation cost and/or those belonging to conversations with longer inactive periods are preferentially evicted.

\paragraph{Estimating the recomputation cost.}
In order to calculate the retention value of a chunk, we need to estimate its recomputation cost.  In particular, we view the cost to recompute the embedding of a chunk as the sum of recomputing
the LLM model's attention operation and the rest of the non-attention operation: $Cost(s, l) = Cost_{attention}(s, l) + Cost_{other}(s)$, where $s$ is the chunk size and $l$ is the size of the context to which the chunk ``attends'' for the attention operation.
The cost of non-attention computation cost ($Cost_{other}(s)$) consists of linear layers, layer normalization,
non-linear activation, etc., and therefore is independent of the context
size. On the contrary, attention requires accessing and performing computation
with all $l$ context tokens. Since the eviction decisions are made for fixed-size chunks of 32 tokens,
we can simplify the cost function to become $Cost(l) = Cost_{attention}(l) + c$ where $Cost_{attention}(l)$ is the cost of performing attention operation for a chunk of 32 tokens with context length $l$, and $c$ is a constant capturing the cost of non-attention computation.
We perform offline profiling to estimate $c$ as well as $Cost_{attention}(l)$ with varying context sizes. Since it's not feasible to profile all possible context sizes, we profile context sizes that are powers of 2 and use the measured values to interpolate the cost for other context sizes.

\subsubsection{Ahead-of-the-time swapping}
Since \system tries to preserve \kvtokens in the GPU for reuse by a later request in the same conversation, the scheduler does not immediately release a request's
GPU cache slots as soon as it finishes, unlike existing systems~\cite{vllm}.
Instead of waiting until the GPU cache has run out, the scheduler asks the
worker to copy (aka swap out) selected \kvtokens to the CPU if less than a
threshold (e.g. 25\%) of the GPU cache slots are available.  The corresponding
GPU memory is reclaimed in a lazy manner and is not immediately released until
the scheduler later decides to allocate the same slots to another conversation.

When the CPU cache runs out of space, the same eviction policy (\S\ref{subsubsec:evict}) is used to
decide which \kvtokens to drop.  Performing ahead-of-the-time swapping allows
the scheduler to overlap cache eviction with GPU computation, thus fully hiding the latency of swapping.

\subsubsection{Pipelined \kvcache recovery.}
The scheduler does not wait for a request's \kvtokens to be fully swapped in from the CPU before handing it off to the worker for execution. Rather, we follow the pipelined  approach~\cite{pipeswitch} to overlap computation with data transfer. Specifically, we exploit the fact that an LLM model has many layers and each layer's \kvtoken is only used in this layer's self-attention calculation. Instead of waiting for all
layers to finish data transfer before starting the execution, we initiate the transfer layer by layer and start model computation at the same time. The worker uses GPU events to preserve data dependency: it only starts a layer's self-attention kernel once that layer's \kvtokens have been fully copied to the GPU. Pipelining transfer with computation allows us to hide the swap-in latency.

\subsubsection{Handling dropped tokens.}
\label{pensieve:subsubsec:sol:recompute}

\begin{figure*}[tbp]
\centering
\includegraphics[width=0.75\textwidth]{\PensievePath/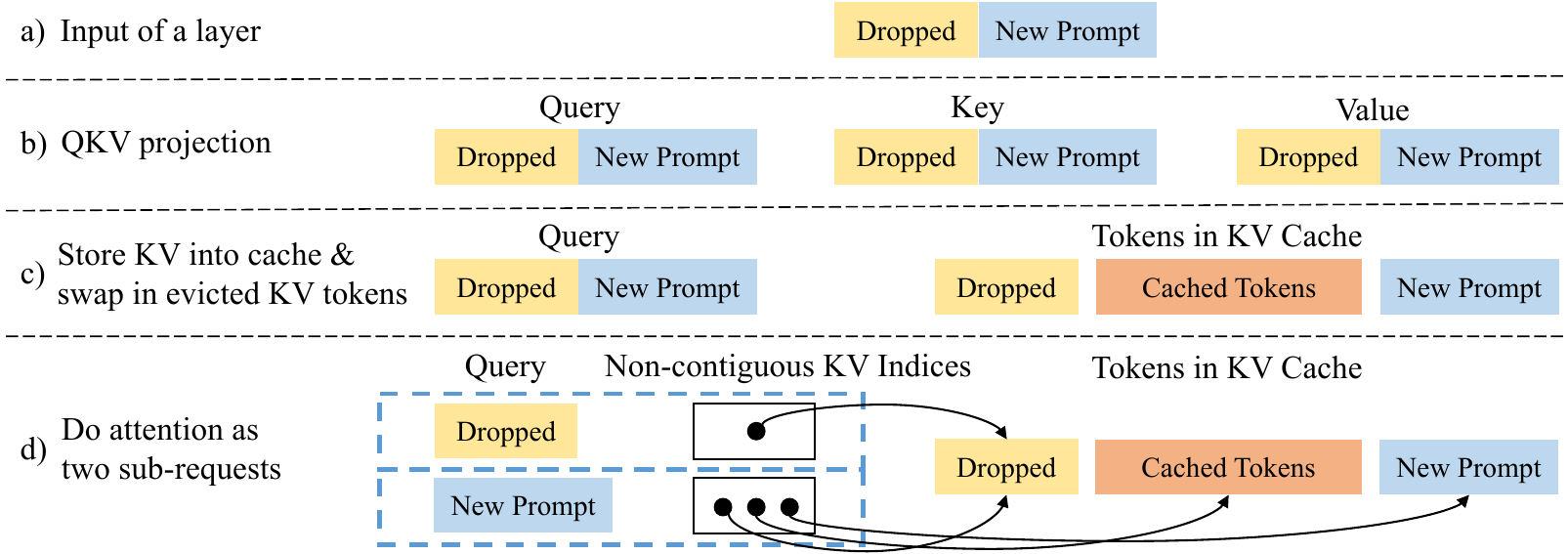}
\caption{\capsize{How \system handles dropped tokens. (a) concatenate tokens of
the dropped context and new prompt as the input. (b) Calculate projections $Q$,
$K$, $V$ (c) Update \kvcache using newly calculated $K$, $V$; swap in \kvtokens
previously evicted to the CPU. (d) Do attention for a batch of two sub-requests:
one for the dropped context to attend to itself, and the other for the new
prompt to attend to the entire context.}}
\label{pensieve:fig:sol:compute_as_both_ends}
\end{figure*}

If a request has some of its \kvtokens dropped due to CPU memory pressure, we
resort to recomputation to handle such dropped tokens.
Figure~\ref{pensieve:fig:challenge:recompute_drop} shows the whereabouts of a typical new request's \kvtokens. As we always evict cached tokens from the leading end of a conversation, the GPU cache generally
holds the request's latest tokens and the CPU cache holds the middle, while the
earlier ones may have been dropped.

The scheduler swaps in those tokens cached in the CPU. For dropped tokens, the
scheduler will fetch their corresponding raw text tokens from the conversation
history saved in a persistent store (Figure~\ref{pensieve:fig:system}).  These
retrieved raw tokens will be merged into (i.e. prepended to) the new request's
prompt and become part of the batch's input tokens, as shown in
Figure~\ref{pensieve:fig:sol:compute_as_both_ends} (step a). In the prefill phase, the
embedding of dropped tokens and new prompt tokens are concatenated together as
they are processed by successive model layers. At each Transformer layer, the Query, Key,
and Value tensors are computed (step b). Key and Value are stored in \kvcache, and
\system maintains the KV locations for the entire conversation context
including previously cached tokens (step c), which can then be used to perform attention.

However, as discussed in \S\ref{pensieve:sec:challenge:challenge}, the challenge that comes with dropping leading tokens is that tokens in query tensor correspond to two disconnected ranges in the context, while all existing
attention kernels assume that the Query tensor region is consecutive. To address this,
we treat these two ranges as two sub-requests that happen to share portions of the underlying context. Each row in Figure~\ref{pensieve:fig:sol:compute_as_both_ends} (step d)
represents the Query tensor and its corresponding KV context locations of each sub-request of the
original request. As our \multiattn GPU kernel design (\S\ref{pensieve:system:multiattn}) can support Query tensors of variable
lengths for different requests in the batch and also accept non-contiguous \kvtoken locations, \system only needs to update auxiliary data
structures and no memory copy is incurred when processing the sub-requests.

\subsubsection{Suspending requests during \gen}
Despite ahead-of-time eviction, the scheduler might still encounter scenarios when the \gen phase runs out of GPU cache since a request's decoding length is not known a priori. In this situation, the scheduler suspends some requests' execution by taking them out of the current batch, swaps out their corresponding \kvtokens to the CPU, and puts them back in the waiting queue. It chooses which request to suspend according to the descending order of their arrival time.
As suspension causes increased latency (due to waiting for the swap-out), we try to avoid it by conservatively reserving 10\% of GPU cache slots for the execution of existing requests that are in the \gen phase.
In other words, the scheduler stops adding new requests into the running batch unless there are more than 10\% free GPU cache slots.

\begin{figure}[tbp]
\centering
\includegraphics[width=\columnwidth]{\PensievePath/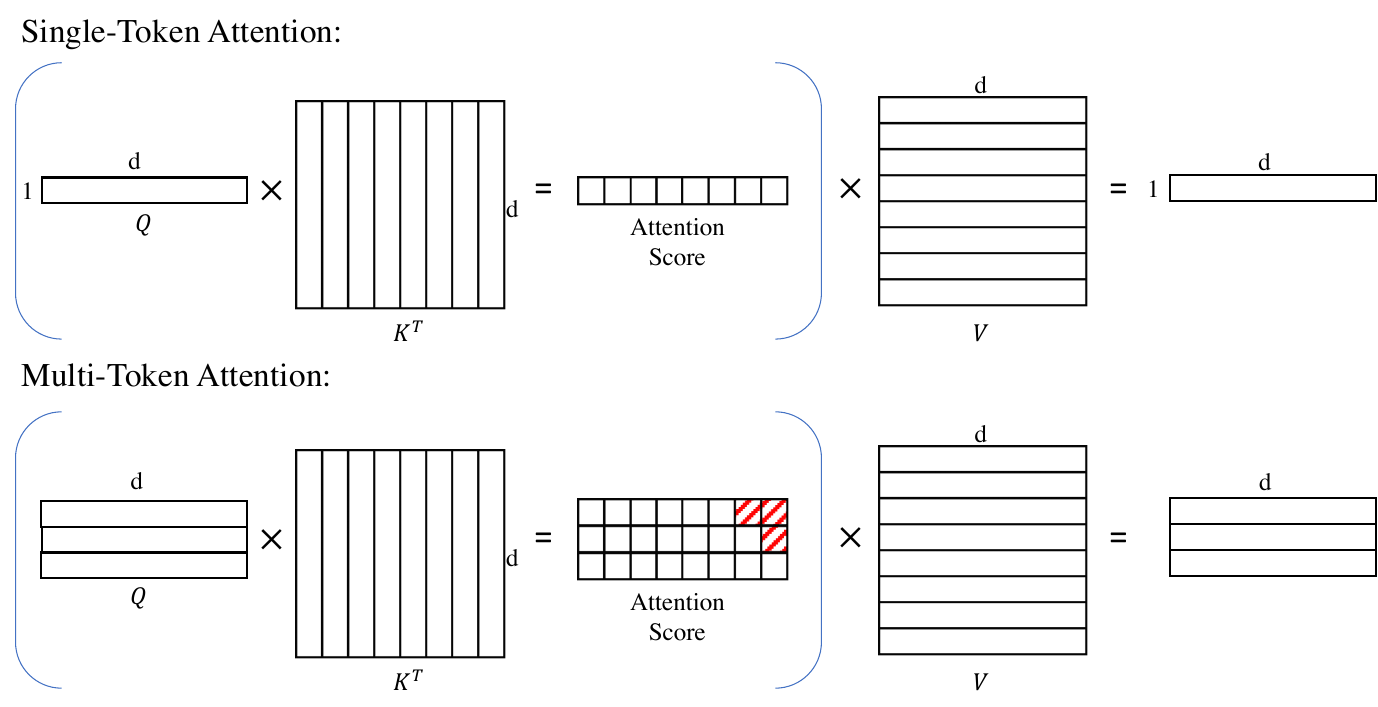}
\caption{\capsize{\Singleattn vs \multiattn, softmax and rescaling omitted. $d$ is the
dimension of the embedding vector. The red shaded area in the \multiattn score
represents causal masking. $K$ and $V$ are logically contiguous, but physically
placed at random locations when paged memory management is used.}}
\label{pensieve:fig:sol:single_vs_multi_query}
\end{figure}

\subsection{\Multiattn for non-contiguous cache}
\label{pensieve:system:multiattn}

How to combine existing \kvtokens in the GPU cache with those just swapped in from the CPU? The most naive solution is to allocate a contiguous memory region in the GPU to hold them both.  However, doing so would incur expensive memory copying, since \kvtokens are large.  A more promising solution is to allocate separate space only for those swapped-in \kvtokens and to design an attention kernel implementation that can handle non-contiguous memory in its \kvcache.

vLLM's PagedAttention kernel can handle non-contiguous \kvcache in the \gen
phase~\cite{vllm}.  However, for the \init phase, it still uses existing kernels
which require all \kvtokens to reside in contiguous memory.   We cannot use
PagedAttention because it assumes each request in a batch has exactly one input
token, which is the case for \gen but not \init. Thus, we refer to
PagedAttention as a \singleattn kernel because it computes the attention scores
between a single input token's query representation ({\sf Q}) and the \kvcache.
We need to build a \multiattn kernel that supports performing
attention between the query representations ({\sf Q}) of multiple input tokens
per request and the \kvcache over non-contiguous memory.

Our new kernel is similar to that of vLLM to the extent that both kernels need
to support loading \kvcache from non-contiguous GPU global memory into on-chip
shared memory. Their main difference is illustrated in
Figure~\ref{pensieve:fig:sol:single_vs_multi_query}. As vLLM's kernel performs
attention between a request's single input token and the existing \kvtokens, its
underlying computation can be described as two matrix-vector multiplication
operations\footnote{The batched version of PagedAttention performs batched
matrix-vector multiplications.}, as shown in
Figure~\ref{pensieve:fig:sol:single_vs_multi_query}. In contrast, our kernel
handles multiple input tokens for each request, computing attention scores
between all pairs of input tokens and the conversation's existing \kvtokens.
Therefore, its underlying computation can be described as two matrix-matrix
multiplication operations, and the batched version of our kernel performs
batched matrix-matrix multiplications.  Like vLLM, we fuse the two
multiplication operations according to~\cite{flash}. Because of the additional
dimension in the {\sf Q} tensor, our kernel has more parallelization and tiling
opportunities on the GPU.
However, care must be taken to handle the new challenge that Q is a ragged-sized tensor as different requests
in the batch have different numbers of input tokens.

When multiple input tokens of a request are handled together in a kernel, we need to apply causal masking so that an earlier input token does not ``attend'' to later tokens.  This requires
setting a corresponding upper triangular part of the attention score matrix to 0, as shown by the red shaded area in Figure~\ref{pensieve:fig:sol:single_vs_multi_query}. Causal masking is not needed for the \singleattn kernel since it only processes one input token.
We fuse the causal masking operation inside the \multiattn kernel to avoid materializing the intermediate attention score matrix~\cite{flash}.

Matrix-matrix multiplication kernels are much more complex than those for matrix-vector multiplication because they typically use sophisticated algorithms to extract additional data reuse opportunities with GPU's on-chip shared memory and to leverage GPU's tensor core primitives. Therefore, instead of trying to extend vLLM's PagedAttention kernel for \multiattn, we base our implementation on an existing \multiattn kernel from PyTorch and extend it to handle non-contiguous \kvcache.  Our kernel uses the high-performance thread-block level matrix-matrix multiplication provided by NVIDIA Cutlass template library~\cite{cutlass}.

\subsubsection{Unifying the \init and \gen phase}
\label{subsubsec:unify}

As discussed in \S\ref{subsec:batchscheduler}, in \system's batch formation, new
requests in their \init phase can be grouped with existing requests in their
\gen phase.  This is enabled by our \multiattn kernel because \singleattn
performed in the \gen phase can be treated as a special case of \multiattn with
query size equal to 1.

More concretely, \system's scheduler concatenates the input tokens to be processed from all requests, regardless of whether they correspond to prompts of new requests or the last step's output token from existing requests. Some auxiliary data structure is maintained to keep track of each request's corresponding region.  During execution by the worker, these batched input tokens are fed through linear layers to generate each token's {\sf QKV}
representations. Newly generated \kvtokens are stored in the allocated slots in the GPU's \kvcache. Then the worker
applies our \multiattn kernel to produce the output tokens for all requests.

\subsubsection{Multi-GPU support}
\label{subsubsec:multi-gpu}
For large LLM models that do not fit into a single GPU, we adopted Tensor
Parallelization, as implemented in Megatron-LM~\cite{megatron}, to partition
large models across multiple GPUs. The \kvcache in \system is also partitioned
accordingly. Each model partition is managed by a worker process which has its
own GPU and CPU cache partition to store the attention state partition assigned
to this worker. Since the model partitioning occurs along the feature dimension
of KV cache, it does not affect the eviction policy that decides which token to
migrate or drop. Therefore, each worker follows the same migration plan to swap
attention states between GPU and CPU memory.

%% file: 5_impl.tex
\section{Implementation}
\label{pensieve:sec:impl}

We have implemented our prototype serving system \system with $\sim$7K
lines of C++/CUDA code. \system manages \kvcache and auxiliary data
structure needed by \multiattn on the GPU, but relies on PyTorch (v2.0.0, CUDA 11.8) C++ front-end
APIs to execute GPU operators in Large Language Models. Based on PyTorch's
fused memory efficient attention, we develop our own fused \multiattn kernel
using NVIDIA Cutlass library to support performing attention with \kvtokens that
reside in non-contiguous memory.

\paragraph{Optimization: Prioritize data retrieval over eviction}
Although PCIe allows full-duplex bidirectional data transfer, in practice, we found that when CPU-to-GPU data transfer is done concurrently with GPU-to-CPU data transfer, there is a significant throughput drop (18-20\%) in both directions. Similar issues
have been reported\footnote{\url{https://forums.developer.nvidia.com/t/data-transfers-are-slower-when-overlapped-than-when-running-sequentially/187542}}.
Since \system performs \kvtoken swap-out ahead of time, there is no urgency to finish the transfer right away. To prevent
eviction from slowing down the swapping in of \pastkvtoken, we set up a waiting mechanism. In particular, if a worker has any ongoing
swap-in task, it waits to perform GPU-to-CPU copy until the swap-in task is done.
Although this conservative approach does not fully utilize the duplex PCIe bandwidth, we find that this optimization performs well and we never run
into the situation that the GPU-to-CPU copying can't catch up.

%% file: 6_eval.tex
\section{Evaluation}

\input{\PensievePath/6_eval_setup}

\input{\PensievePath/6_eval_e2e}

\input{\PensievePath/6_eval_kernel}

\input{\PensievePath/6_eval_unified}

\input{\PensievePath/6_eval_policy}

\input{\PensievePath/6_eval_react}

%% file: 6_eval_setup.tex
\subsection{Experimental Setup}
\label{pensieve:sec:eval:setup}
\paragraph{System Environment.}
We evaluate \system on Azure NC A100 v4 series, which are equipped with up to 4
A100-80GB GPUs, a 24-core AMD EPYC 7003 processor and 220 GB CPU
memory per GPU. For each system evaluated, we configure it to allocate 40 GB
memory on each GPU for \kvcache for a fair
comparison.

\paragraph{Models}
\begin{table}[tbp]
\centering
\resizebox{\columnwidth}{!}{%
\begin{tabular}{c|cccc}
Model       & \opt-13B & \opt-66B & \llama-13B & \llama-70B \\ \hline
\# layer    & 40      & 64      & 40         & 80         \\
\# hidden   & 5120    & 9216    & 5120       & 8192       \\
\# head     & 40      & 72      & 40         & 64         \\
\# KV head  & 40      & 72      & $10^\ast$  & 8          \\
Head size   & 128     & 128     & 128        & 128        \\
\# GPU      & 1       & 4       & 1          & 4
\end{tabular}
}
\caption{\capsize{Hyper-paramaters for \opt and \llama models.}}
\label{pensieve:table:eval:models}
\end{table}
We use two open-source models: \opt~\cite{opt} and \llama~\cite{llama2}.
\opt has an almost identical model architecture to GPT-3~\cite{gpt3} while
\llama is a more recent model that
employs more advanced model features like rotary embedding, RMS
Layernorm~\cite{rmsnorm}, SiLU, etc. Notably, \llama follows the trend of
adopting \gqa (GQA)~\cite{gqa} which divides query heads into
groups so that only one KV head is used within each group. GQA
significantly reduces the memory consumption of \kvtokens, allowing \system to
store more \pastkvtokens.

We evaluate two different sizes for each model: a small one on a single GPU,
and a large one partitioned onto 4 GPUs using Tensor Parallelization as done in
Megatron-LM (\S\ref{subsubsec:multi-gpu}).  Detailed model hyper-parameters can be found in
Table~\ref{pensieve:table:eval:models}.  By default, \llama only uses GQA for
models with over 70 billion parameters. To demonstrate \system's effectiveness
when used with GQA, we changed the number of KV heads of \llama-13B from 40 to
10. In all experiments, the 16-bit half-precision float format is used for both
model parameters and intermediate hidden representations.

\paragraph{Dataset}

We evaluate \system on two multi-turn conversation datasets: \sharegpt and \ultrachat. \sharegpt~\cite{sharegpt} is a real-world dataset containing
user-shared ChatGPT conversations. \ultrachat~\cite{ding2023enhancing} is a recent large-scale synthetic dataset for multi-turn dialogue: it uses separate LLMs to simulate the interaction between a user
and the chatbot assistant.  Table~\ref{table:eval:dataset} shows the statistics of
both datasets.  In our experiments, we limited the maximum context size to 16384
tokens and dropped 0.57\% of the conversations in \sharegpt dataset that exceed
this limit.

\begin{table}[tbp]
\small
\centering
\begin{tabular}{c|rr}
                      & \sharegpt & \ultrachat\\ \hline
\# conversations      & 48,159  &  1,468,352 \\
Mean \# of turns      & 5.56  & 3.86  \\
Mean request input length  & 37.77 & 51.78 \\
Mean request output length & 204.58 & 257.81
\end{tabular}
\vspace{0.1in}
\caption{\capsize{Dataset statistics}}
\label{table:eval:dataset}
\end{table}

\paragraph{Workload}
Since the datasets do not provide timestamps for each user request, we simulate a request's arrival time by sampling from a Poisson distribution under different request rates. We maintain the causal dependency for requests
belonging to the same conversation: a new user prompt is only sent to the system
after the response to the conversation's previous request has been received.
Additionally, we also simulate user think time, aka the time taken for users to generate the next conversation turn, by sampling from an exponential distribution with varying mean.

\begin{figure*}[htbp]
    \centering
    \begin{subfigure}{\columnwidth}
        \centering
        \includegraphics[width=0.95\columnwidth]{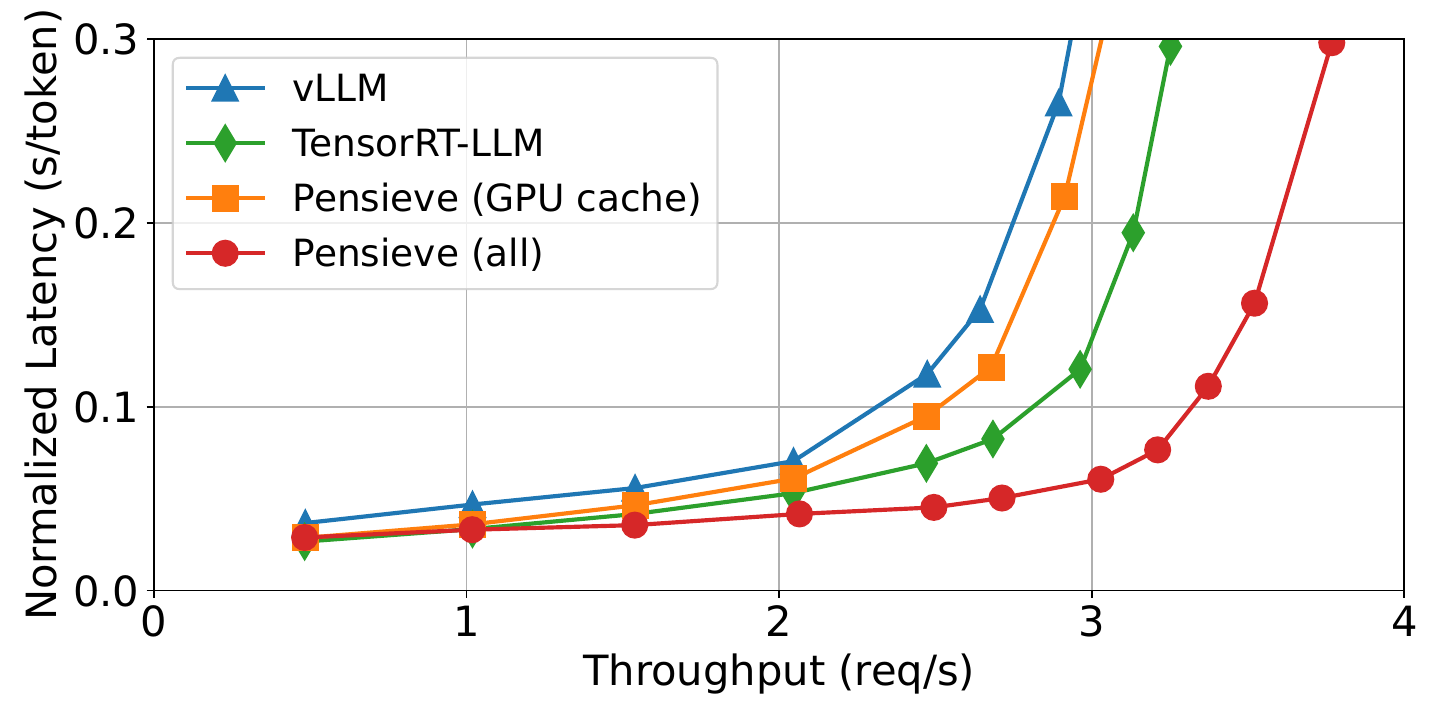}
        \caption{\capsize{\opt-13B, \sharegpt}}
        \label{pensieve:fig:eval:e2e:opt13b}
    \end{subfigure}
    \begin{subfigure}{\columnwidth}
        \centering
        \includegraphics[width=0.95\columnwidth]{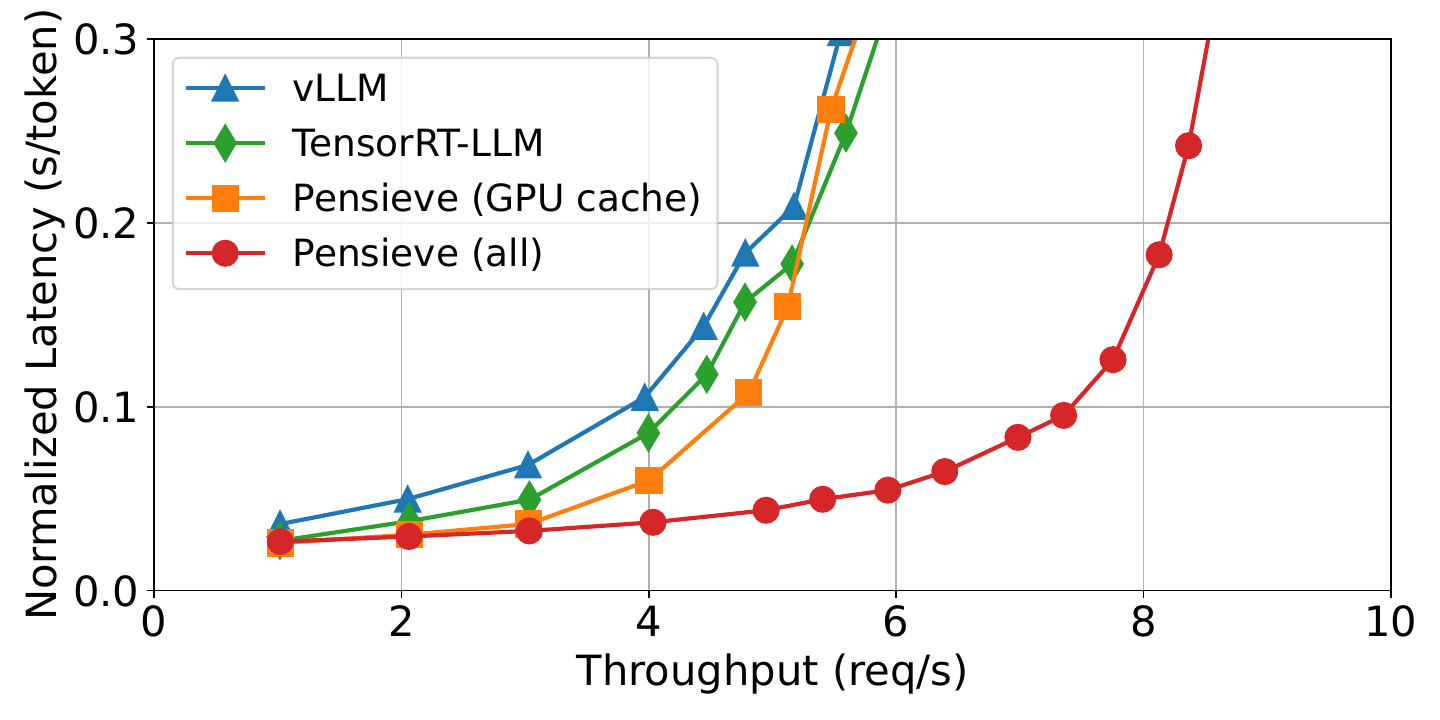}
        \caption{\capsize{\llama-13B, \sharegpt}}
        \label{pensieve:fig:eval:e2e:llama13b}
    \end{subfigure}
    \begin{subfigure}{\columnwidth}
        \centering
        \includegraphics[width=0.95\columnwidth]{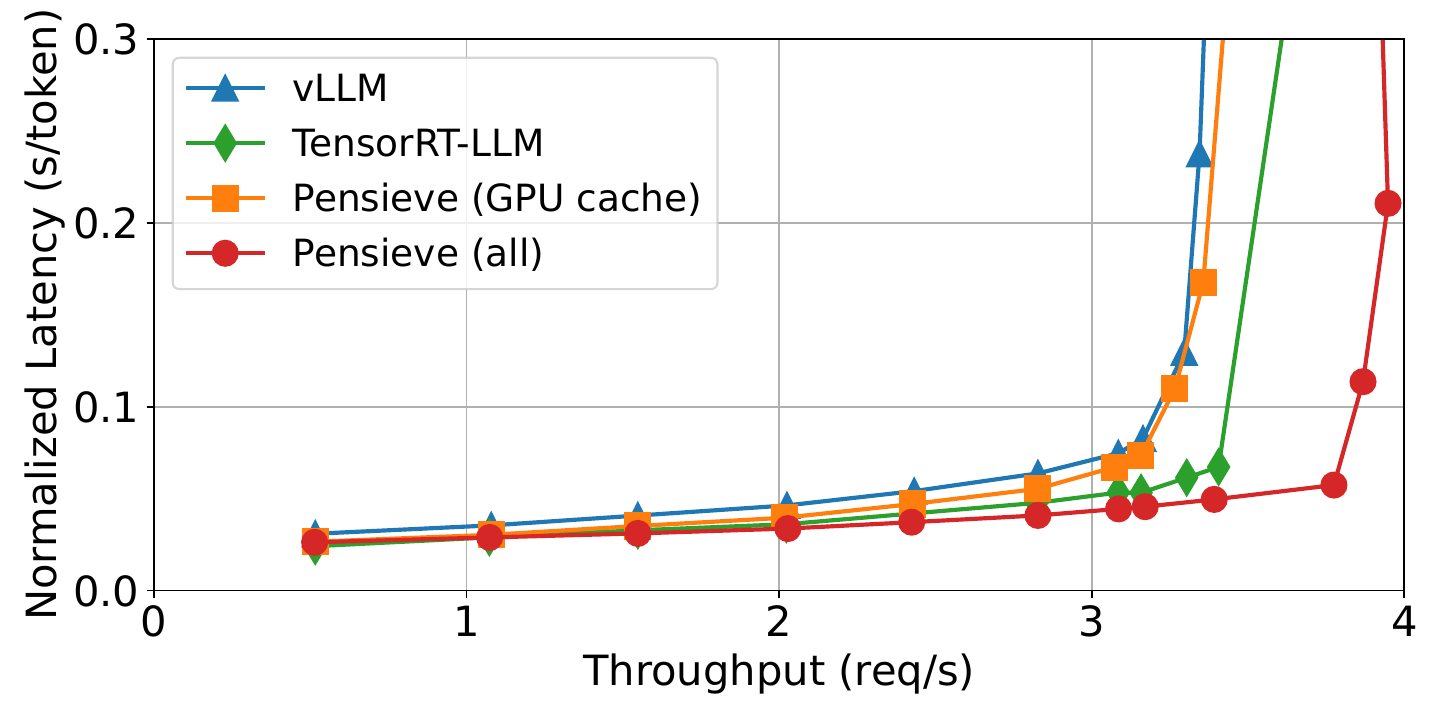}
        \caption{\capsize{\opt-13B, \ultrachat}}
        \label{pensieve:fig:eval:e2e:opt13b_ultrachat}
    \end{subfigure}
    \begin{subfigure}{\columnwidth}
        \centering
        \includegraphics[width=0.95\columnwidth]{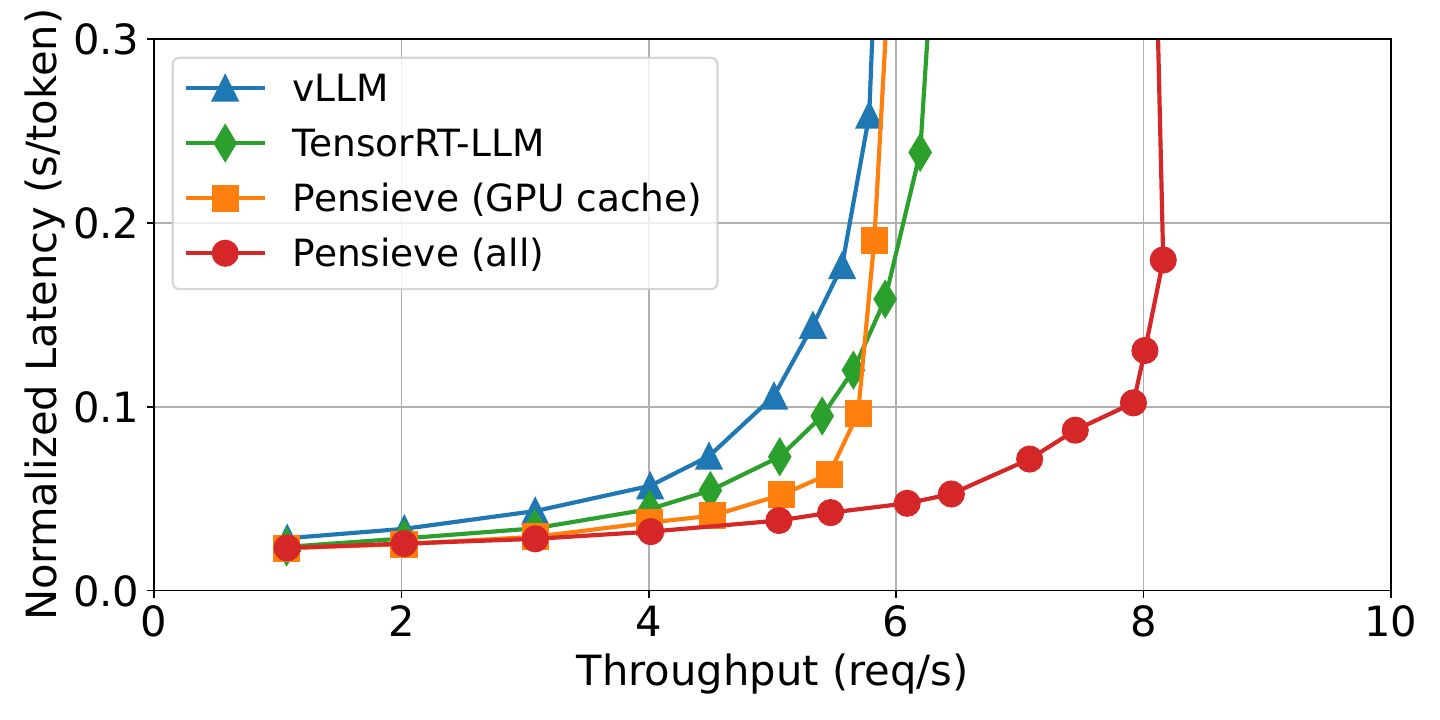}
        \caption{\capsize{\llama-13B, \ultrachat}}
        \label{pensieve:fig:eval:e2e:llama13b_ultrachat}
    \end{subfigure}
    \caption{\capsize{LLM serving performance on 1 GPU.}}
    \label{pensieve:fig:eval:e2e:1gpu}
\end{figure*}

\paragraph{Baselines}
We evaluate \system against two state-of-the-art serving systems:
\vllm~\cite{vllm} (v0.2.0) and TensorRT-LLM~\cite{trtllm} (v0.12.0).  Both of
them use a stateless serving API. For each request, the new user prompt is appended
to the history and then processed as an input request. \vllm uses PyTorch as its execution backend. By contrast, TensorRT-LLM compiles and optimizes the model using graph rewriting optimizations such as operator fusion, and executes the optimized model using the TensorRT Runtime. Since \vllm is an execution engine without a serving
loop that drives the engine to process incoming requests, we implement such a driver
that adds newly arrived requests into \vllm job queue and invokes the engine
execution until all requests are fully processed.

We also experiment with a variant of \system called \system (GPU cache) that simply drops evicted tokens from the GPU instead of swapping them out to the CPU.  This variant is used to
examine the effectiveness of CPU caching in \system.

\paragraph{Performance Metric}
\system is designed to optimize both peak throughput and latency of
serving LLM in conversational scenarios. Following prior work~\cite{orca,vllm},
we measure the achieved serving throughput and 90-percentile normalized
latency, which is calculated as the end-to-end request serving
latency divided by the number of output tokens.

%% file: 6_eval_e2e.tex
\subsection{End-to-End Serving Performance}
Figure~\ref{pensieve:fig:eval:e2e:1gpu} shows the normalized latency vs
throughput for \opt-13B and \llama-13B served with a single A100-80GB GPU. The
normalized latency is calculated as the mean of each request's end-to-end
latency divided by its output length, as done in~\cite{orca,vllm}.  On the
\sharegpt dataset shown in Figure~\ref{pensieve:fig:eval:e2e:1gpu}(a) and (b),
\system's throughput is $1.36\times$ of \vllm and $1.14\times$ of \trtllm for
serving \opt-13B at 120 ms per token latency, and
$1.70\times$ of \vllm and $1.58\times$ of \trtllm  for
serving \llama-13B at 180 ms per token latency. On the \ultrachat dataset shown in
Figure~\ref{pensieve:fig:eval:e2e:1gpu}(c) and (d), \system's throughput is $1.17\times$ and
$1.14\times$ of \vllm and \trtllm respectively for serving \opt-13B at 120 ms per token latency,
$1.58\times$ and $1.47\times$ for serving \llama-13B at 100 ms per token latency. \system has more
throughput gains on \sharegpt than \ultrachat because the real world dataset
\sharegpt has more conversation turns (Table~\ref{table:eval:dataset}), which
makes saving \pastkvtokens more beneficial.

In Figure~\ref{pensieve:fig:eval:e2e:1gpu}, \trtllm outperforms
\vllm consistently.
It even outperforms \system (GPU cache) in
Figure~\ref{pensieve:fig:eval:e2e:1gpu}(a) and (c).
This is because both \system and \vllm use PyTorch API to execute the model
while \trtllm performs optimizations like
graph rewriting and operator fusion during an offline compiling stage, and then
executes the optimized model using the TensorRT Runtime.  However, by avoiding recomputing
\pastkvtokens for continuing conversations, \system achieves significant throughput gain over \trtllm.

\system's performance gain is more significant for \llama-13B than for \opt-13B because our version of \llama-13B uses \gqa
with group size 4 (i.e. every four query heads share a single key-value head). Consequently, the amount of memory required to store
\pastkvtokens is reduced by $4\times$, thus allowing \system to store more \pastkvtokens and better avoid recomputation.

When \system is used without CPU cache, i.e. \system (GPU cache), it may still
reduces latency because baseline systems always recompute \pastkvtokens for
requests from a returning conversation, and thus, the number of tokens processed
in the prefill phase is on average larger. But under relatively higher request
rates, GPU cache is quickly exhausted, and \system (GPU cache) also resorts to
recomputing \pastkvtokens from scratch.

\subsection{Multi-GPU Serving Performance}
\begin{figure}[tbp]
    \centering
     \begin{subfigure}{\columnwidth}
         \centering
         \includegraphics[width=0.95\columnwidth]{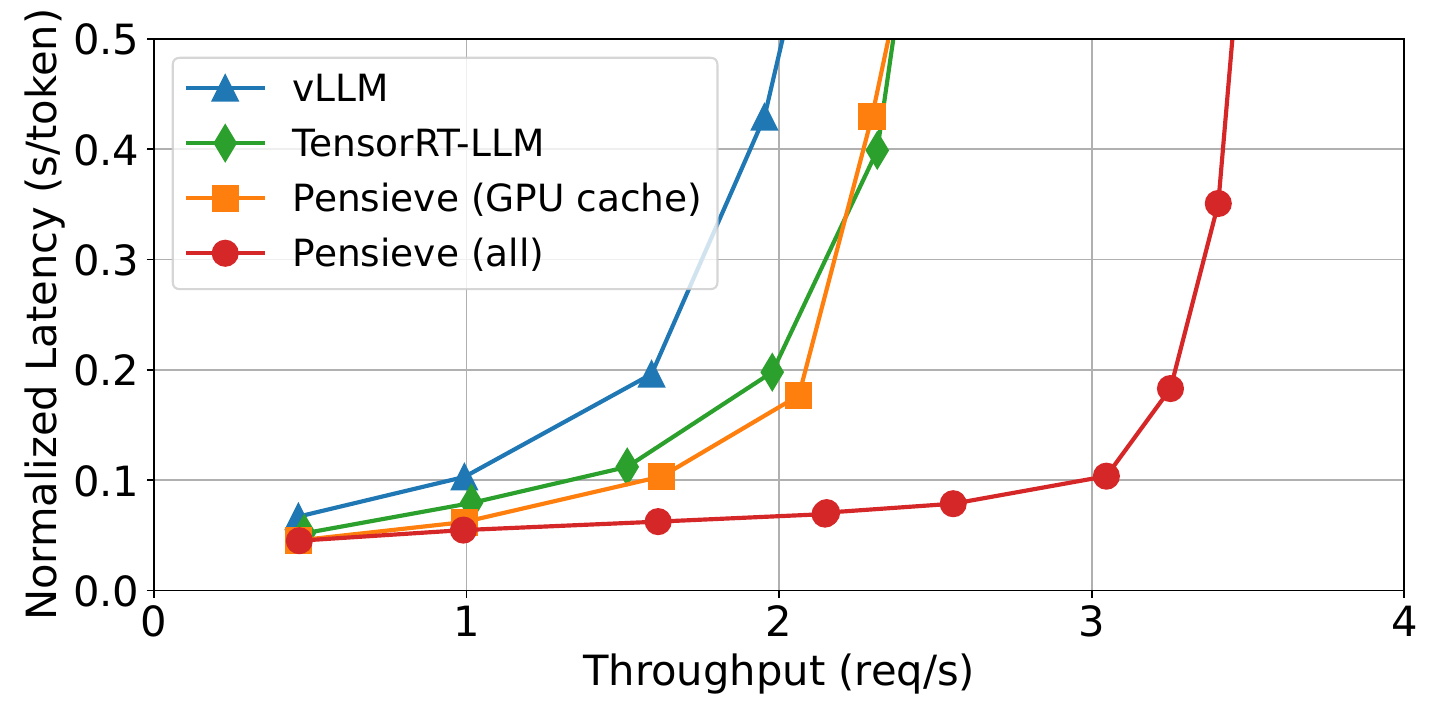}
         \caption{\capsize{\opt-66B}}
         \label{pensieve:fig:eval:e2e:opt66b}
     \end{subfigure}
     \begin{subfigure}{\columnwidth}
         \centering
         \includegraphics[width=0.95\columnwidth]{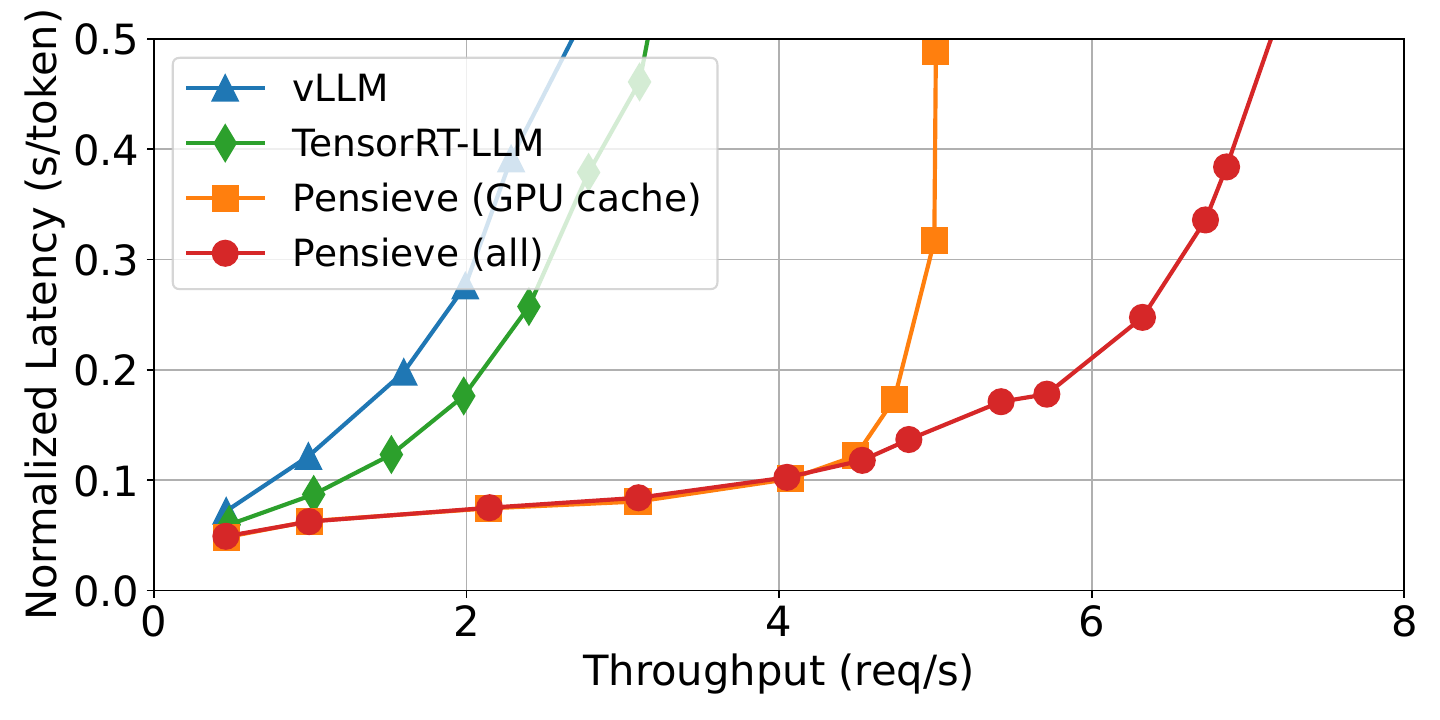}
         \caption{\capsize{\llama-70B}}
         \label{pensieve:fig:eval:e2e:llama70b}
     \end{subfigure}
     \caption{\capsize{LLM serving performance on 4 GPUs for \sharegpt.}}
     \label{pensieve:fig:eval:e2e:4gpu}
\end{figure}

Figure~\ref{pensieve:fig:eval:e2e:4gpu} shows the performance of \system for
larger models, \opt-66B and \llama-70B, when run on four GPUs using the ShareGPT dataset.  Larger models
amplify \system's advantage over the baselines because the amount of computation grows faster than the memory usage of \kvtokens. For example, from \opt-13B to
\opt-66B, the model parameter size and computation amount increase by more than
$5\times$, while the \kvcache size (\# layer $\times$ \# hidden $\times$ 2) only increases by $2.88\times$
(Table~\ref{pensieve:table:eval:models}). Since the number of GPUs and CPU
memory are usually scaled linearly with the model size, \system can store more
\pastkvtokens in its \kvcache.

\system achieves $2.04\times$ the throughput of \vllm and $1.64\times$ of \trtllm for
serving \opt-66B at 200 ms per token latency, and $3.0\times$ of \vllm and $2.47\times$ of \trtllm for serving \llama-70B at 400 ms per token latency.
The improvement is more significant on \llama-70B because it
uses \gqa with group size 8, which reduces the memory requirement for
\pastkvtokens by $8\times$. This much-reduced memory consumption in \kvtokens also
significantly benefits the throughput of \system (GPU cache), as shown in
Figure~\ref{pensieve:fig:eval:e2e:4gpu}(b).

%% file: 6_eval_kernel.tex
\subsection{Performance of \Multiattn Kernel}
\label{pensieve:sec:eval:kernel}

\begin{figure}[tbp]
    \centering
    \includegraphics[width=0.9\columnwidth]{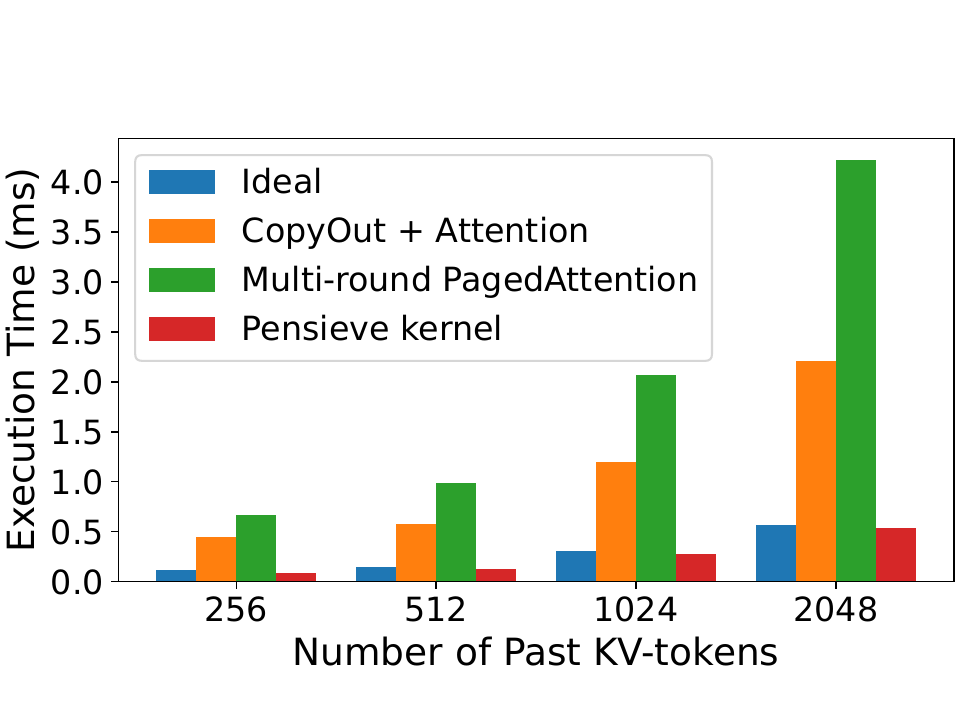}
    \caption{\capsize{Execution time of \system's \multiattn kernel over
    non-contiguous context memory with different context sizes. Measured with
    batch size 32 and query size 8.}}
    \label{pensieve:fig:eval:kernel}
\end{figure}

Figure~\ref{pensieve:fig:eval:kernel} shows the performance of \system's \multiattn kernel compared to alternative implementations. In this microbenchmark, we measure the latency of the attention operator for a batch of 32 requests each with a prompt of 8 query tokens and different numbers of \pastkvtokens stored in non-contiguous GPU memory. As
described in \S\ref{pensieve:sec:challenge:motivation}, existing attention kernels are not directly applicable. We compare
against two straw-man implementations: (1) ``CopyOut+Attention'' allocates additional contiguous GPU memory to copy \pastkvtokens into and
then invokes an existing fused attention kernel (orange bar), and 2) ``Multi-round PagedAttention'' invokes multiple rounds of \vllm's single-query PagedAttention to
process one token from the prompt at a time (green bar). We also show the performance of the ideal situation which assumes that
\pastkvtokens are stored in contiguous memory space (blue bar).  Figure~\ref{pensieve:fig:eval:kernel} shows that both straw-man
solutions add significant overhead compared to the ideal performance. Copying to contiguous memory incurs cost proportional to the number of \pastkvtokens. Applying multiple
rounds of PagedAttention, on the other hand, gives up parallelization opportunities on prompt tokens, resulting in
execution time linear to the number of tokens in the prompt. \system's kernel matches the ideal baseline.  In fact, it has slightly better performance
because we offload auxiliary data computing (like calculating the cumulative sum for the sequence length of a batch) to the CPU.
Since each transformer layer in the model shares the same caching plan, these auxiliary data can be reused by all
layers.

%% file: 6_eval_unified.tex
\subsection{Effect of Unified Scheduling}
\begin{figure}
    \centering
    \includegraphics[width=0.95\columnwidth]{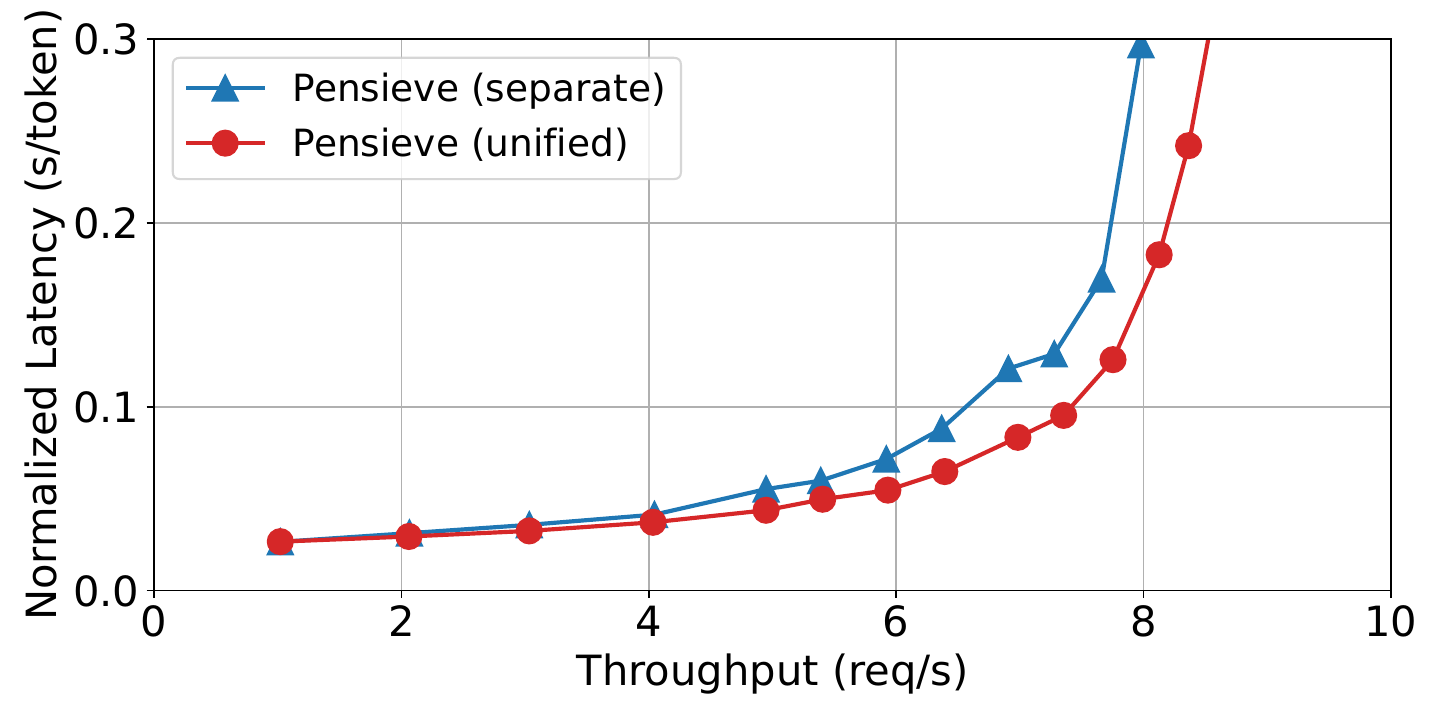}
    \caption{\capsize{Performance of serving \llama-13B with and without unified scheduling on \sharegpt dataset.}}
    \label{pensieve:fig:eval:scheduler}
\end{figure}
We evaluate whether using a unified scheduler for both
\init and \gen phases is beneficial for performance.
Figure~\ref{pensieve:fig:eval:scheduler} shows the performance of \system
with and without unified scheduling for \llama-13B on the \sharegpt dataset. Compared to processing \init and \gen phases
separately, unifying them into a single execution step avoids having to execute the
\init phase with a small number of requests. As a result, \system with unified scheduling achieves
better throughput and latency.

%% file: 6_eval_policy.tex
\subsection{Effect of the Eviction Policy}
\begin{figure}[tbp]
    \centering
    \includegraphics[width=0.95\columnwidth]{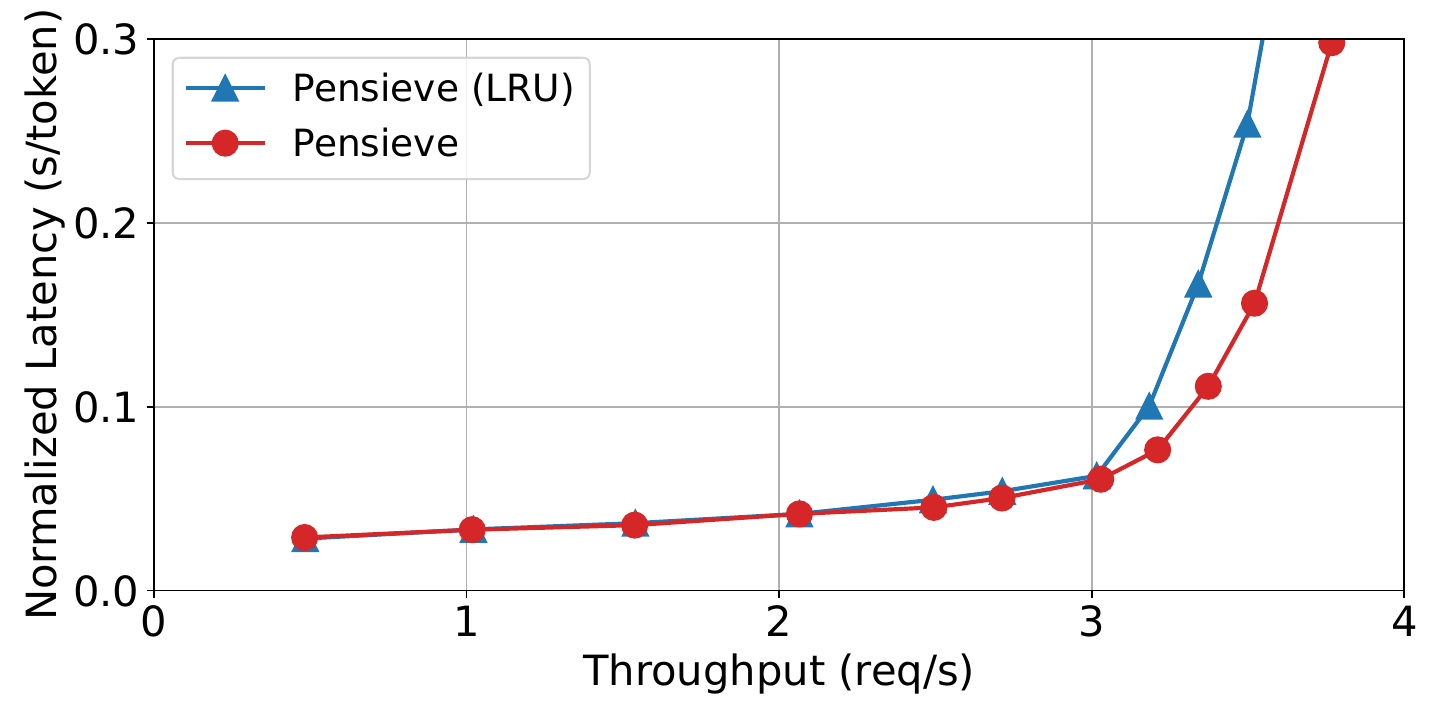}
    \caption{\capsize{Performance of serving \opt-13B with different eviction policies on \sharegpt dataset.}}
    \label{pensieve:fig:eval:eviction}
\end{figure}

We compare \system's caching policy against the classic LRU
policy. We use \opt-13B as the workload for this evaluation.
Figure~\ref{pensieve:fig:eval:eviction} shows that both policies exhibit similar
performance until the workload approaches 3 requests per second, beyond which
\system's eviction policy outperforms LRU.
After analyzing the execution traces, we find that
both policies have less than 80\% cache hit rate, however,
\system's policy has up to 4.4 percentage points higher CPU cache hit rate than LRU. On
average, \system's policy reduces the number of recomputed \kvtokens by up
to 14.6\%.

%% file: 6_eval_react.tex
\subsection{Impact of User Think Time}
\label{pensieve:sec:eval:react}
\begin{figure}[tbp]
    \centering
    \includegraphics[width=0.95\columnwidth]{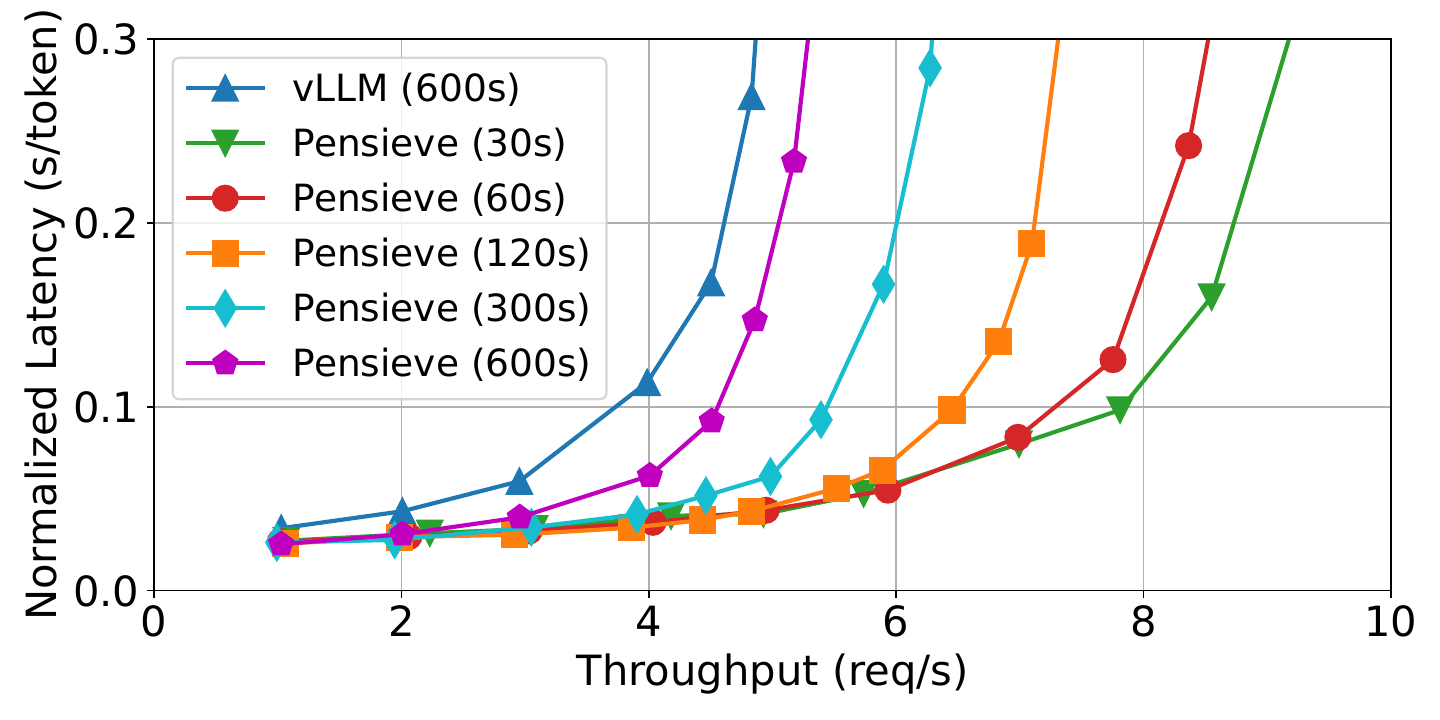}
    \caption{\capsize{Performance of serving \llama-13B with different average
    user reaction time on \sharegpt dataset.}}
    \label{pensieve:fig:eval:react_time}
\end{figure}

Our experiments so far use the average user think time of 60 seconds. Figure~\ref{pensieve:fig:eval:react_time} evaluates the impact of different average user think times on the performance of \system using
\llama-13B. Additionally, we also show \vllm
with 600-second think time as a comparison point.
As seen in Figure~\ref{pensieve:fig:eval:react_time},
the throughput of \system decreases as the average user think time increases, causing \pastkvtokens to drop from the cache at a higher rate. Even as user think time increases to 600
seconds, \system still achieves better latency and throughput compared to \vllm, although the performance gap becomes smaller than that for smaller think times.

%% file: 7_rel.tex
\section{Related Works}
\label{pensieve:sec:related}

\paragraph{LLM inference and serving systems.}
Many systems have been recently developed to serve language models with better
performance including vLLM~\cite{vllm}, \ORCA~\cite{orca},
TensorRT-LLM~\cite{trtllm}, FasterTransformer~\cite{fastertransformer},
LightSeq~\cite{lightseq}, DeepSpeed\xspace\cite{deepspeedinference} and
~\cite{pope2023efficiently}.  These systems investigate different performance
improvement opportunities than \system. The wide range of proposed techniques
include incremental decoding~\cite{fastertransformer}, iteration-level
batching~\cite{orca, batchmaker}, supporting paged KV cache over non-contiguous
GPU memory~\cite{vllm,trtllm}, finding an efficient multi-device partitioning
plan~\cite{pope2023efficiently}, better load balancing between prefill and
generation to reduce pipeline parallelism
bubbles~\cite{sarathi,deepspeed_fastgen}, kernel fusion~\cite{flash,
flashdecode, memeff}, speculative
decoding~\cite{speculative:chen,speculative:leviathan}, and
quantization~\cite{nagel2021white,gholami2022survey, dettmers2023case,
smoothquant2023}.

ServerlessLLM~\cite{serverlessllm} proposes techniques to do fast model loading
and live migration in order to serve requests in a serverless setup.
DistServe~\cite{distserve} advocates performing the \init and \gen phase of the
same request on separate GPUs to better satisfy SLA. By contrast, \system
performs unified batching of both phases because we aim to optimize throughput
instead of latency SLA.

\begin{footnotesize}
\begin{table*}[]
\centering
{
\begin{tabular}{c|cccccc}
                            & PromptCache~\cite{promptcache} & SGLang~\cite{sglang}                & CachedAttention~\cite{cachedattn}  & RAGCache~\cite{ragcache} & CacheGen~\cite{cachegen}  & Pensieve             \\ \hline
Cache shared across users   & Yes         & Yes                   & No               & Yes      & Yes & No                   \\
Cache storage               & CPU          & GPU                   & CPU \& Disk      & CPU \& GPU & Network  & CPU \& GPU           \\
Eviction granularity             & N/A         & requests & entire conversation history &  documents & N/A & tokens \\
Eviction location preference & N/A & trailing & N/A & trailing & N/A & leading \\
\end{tabular}
}
\caption{\capsize{Comparison of \system with other caching systems for LLM attention states.}}
\label{tab:related}
\end{table*}
\end{footnotesize}

\paragraph{Systems that cache LLM attention states.}
Several systems (Table~\ref{tab:related}), developed concurrently with \system, also cache attention states in LLM serving to reduce computation.

PromptCache~\cite{promptcache}, SGLang~\cite{sglang}, RAGCache~\cite{ragcache} and CachedAttention~\cite{cachedattn} all cache attention states
of shared text segments across different requests.  Among these systems,  CachedAttention targets multi-turn conversations, RAGCache targets retrieval-augmented generation (RAG) while others aim to be general.
PromptCache requires users to define a schema a priori to explicitly specify any reuse opportunities. This requirement is too restrictive since static schemas cannot specify those dynamic prefixes to be cached in multi-turn conversations. SGLang and RAGCache employ a tree-based cache in order to enable prefix sharing across different users or different RAG document prefixes. By contrast, \system and CachedAttention are designed specifically for multi-turn conversation
scenarios where different users are unlikely to share conversational
content\footnote{If the chatbot uses a common system prompt across users, it
can be handled by explicitly designating the system prompt state as reusable.}.
In this setting, there is no need for the extra complexity of a tree-based cache structure. Instead, \system and CachedAttention both focus on managing a
dynamic cache for ongoing conservation histories which grow in size over time.

As Table~\ref{tab:related} shows, the important difference between \system and others (SGLang~\cite{sglang}, RAGCache~\cite{ragcache} and CachedAttention~\cite{cachedattn}) lies in \system's eviction strategy which evicts at the finer-granularity of groups of tokens at the leading end to reduce the cost of recomputation. By contrast, CachedAttention evicts at the granularity of an entire conversation and does not recompute truncated conversation context, while SGLang and RAGCache evict at the granularity of a tree node containing an entire user request or document. SGLang and RAGCache also preferentially evict from the bottom of the tree (aka the trailing end of a prefix) as opposed to the leading end as done in \system.  \system's eviction strategy introduces challenges in
supporting computation in non-sequential regions of request context
(Figure~\ref{pensieve:fig:challenge:recompute_drop}), which is addressed in
\system's multi-token attention kernel
(\S\ref{pensieve:subsubsec:sol:recompute}).

Another recent system, CacheGen~\cite{cachegen}, addresses the problem of
reducing the context-loading delay for long contexts stored in remote network
storage. CacheGen proposes an adaptive compression scheme to accelerate the
transfer of the KV cache over a bandwidth-limited and bursty network.  As
\system only keeps the KV cache in local GPU and CPU memory, CacheGen's
techniques are orthogonal to those of \system. For conversational chatbots,
\system using local cache can already bring significant performance gains.

\paragraph{Non-LLM specific DNN serving systems.}
Systems like TensorFlow Serving~\cite{tfserving}, Clipper~\cite{clipper}, NVIDIA
Triton Inference Server~\cite{triton}, Nexus~\cite{nexus}, and
InferLine~\cite{inferline} serve as scheduling components of a general-purpose
DNN serving system. They are mostly model-agnostic and execution
backend-agnostic and apply general system techniques like batching, caching, and
software pipelining to serve DNN models. Some are also in charge of properly
provisioning compute resources to improve overall cluster efficiency. A few
existing works target model-less serving and provide inference as a service:
they automatically select models to meet the accuracy and latency requirements
of a given user task. For example, INFaaS~\cite{infaas} and Tabi~\cite{tabi}
serve requests with a small model and only re-route to a larger model when the
output confidence score is low.

\paragraph{Techniques addressing the GPU memory limit.}
These include GPU-CPU swapping, recomputation, and unified memory. Most of the
systems described below are not specifically targeted for LLM serving.

{\em GPU-CPU Swapping:}
SwapAdvisor~\cite{swapadvisor} swaps weight and activation tensors for DNN training. It
uses the dataflow graph to determine an optimal plan that involves operator
execution order, memory allocation, and swapping.
Zero-Offload~\cite{zerooffload} offloads optimizer state and gradients to the
CPU during LLM training with a single GPU.
DeepSpeed-ZeroInference~\cite{deepspeedinference} and FlexGen~\cite{flexgen} use
offloading to serve LLM with a weak GPU.  DeepSpeed-ZeroInference offloads
entire model weights to CPU or NVMe memory. FlexGen offloads currently unused
model weights, activation, or \kvcache to CPU memory or disk.  As a result of
frequent data movement and high disk latency, these two systems require a large
batch size to hide the offloading latency. Therefore, they mainly target
latency-insensitive applications. Neither system persists the KV cache across
requests.

{\em Recomputation:}
For DNN training, recomputing activation on the backward pass~\cite{recompute}
is a popular technique used to reduce memory footprint.  The decision for which
tensors to compute is done at the granularity of layers~\cite{recompute}, a
group of operators within a layer (aka modules)~\cite{reducing}, or individual
operators~\cite{checkmate}.  Unlike these works, \system's fine-grained
recomputation is done at the token(chunk)-level, to exploit the insight that
earlier tokens in a sequence incur less recomputation cost due to the causal
nature of attention.

{\em Unified Memory and Direct Host Access:}
Swapping between CPU and GPU memory can also be achieved transparently through
Unified Memory~\cite{unifiedmemory}, which automatically triggers memory page
migration between CPU and GPU using a page fault mechanism.  The Direct Host
Access feature allows a GPU kernel to directly read from CPU memory, thereby
allowing a subset of the threads to execute as soon as their data is ready
without waiting for the entire transfer to finish. It has been used to enable
better overlap of data transfer and kernel computation when swapping in a model
from CPU~\cite{direct} or accessing large graph neural network features on the
CPU~\cite{deepplan}.
For our implementation of \system, we choose not to use Unified Memory nor
Direct Host Access because these mechanisms trigger memory transfer only when
the data is accessed by a GPU kernel and we want to manage data movement
explicitly to prefetch the KV cache.  Furthermore, for Direct Host Access, since
the copied data is not explicitly stored in GPU memory, it has to be loaded from
the CPU again if it is repeatedly used. In our workload, the context cache is
not only used to prefill the prompt tokens but also in every generation step,
which makes Direct Host Access significantly less efficient.

%% file: 8_con.tex
\section{Conclusion}
When serving Large Language
Models for multi-turn conversations, a major inefficiency is due to the recomputation of a conversation's cumulative history context. We
develop \system, a stateful LLM serving system that preserves history embeddings in
a multi-tier GPU-CPU cache. It uses a new GPU attention kernel to perform attention between requests' new multi-token input and their
saved context stored in non-contiguous GPU memory. Experiments show that \system
achieves $1.14$-$1.70\times$ the throughput of baseline systems for small models (OPT-13B, Llama 2-13B) and $1.64$-$3.0\times$ for larger models (OPT-66B, Llama 2-70B).